\documentclass[10pt,twocolumn,letterpaper]{article}

\usepackage[pagenumbers]{cvpr}

\usepackage{needspace}
\usepackage{etoolbox}
\pretocmd{\section}{\needspace{7\baselineskip}}{}{}
\pretocmd{\subsection}{\needspace{4.5\baselineskip}}{}{}

\usepackage{nicematrix}
\usepackage{multirow}
\usepackage{tikz}
\usetikzlibrary{arrows.meta,positioning,fit,backgrounds,calc}
\newcommand{\ours}{PlantWeaver\xspace}

\makeatletter
\newcommand{\para}[1]{%
  \@startsection{paragraph}{4}{\z@}%
    {\glueexpr 3.25ex plus 1ex minus .2ex - 0.0mm\relax}%
    {-1em}%
    {\normalfont\normalsize\bfseries}%
    {#1}%
}
\makeatother

\newcounter{todos}
\AtEndDocument{\ifnum\value{todos}>0 \PackageWarning{TODOS}{There are \arabic{todos} todos left in this paper! Fix them before submitting the paper!} \fi}

\definecolor{cvprblue}{rgb}{0.21,0.49,0.74}
\usepackage[pagebackref,breaklinks,colorlinks,allcolors=cvprblue]{hyperref}

\title{PlantWeaver: Growing Simulation-Ready Plant Populations\\ with Botanically Grounded Agents}

\author{%
Junhao Xing$^{1}$ \quad Risa Shinoda$^{2}$ \quad Yang Yang$^{1}$ \quad Meng-Yu Jennifer Kuo$^{3}$\\
Hiroaki Santo$^{1}$ \quad Shangzhe Wu$^{4}$ \quad Fumio Okura$^{1}$\\[0.25em]
{\normalsize $^{1}$The University of Osaka \quad $^{2}$The University of Tokyo \quad $^{3}$Nara Women's University \quad $^{4}$University of Cambridge}\\[0.05em]
{\normalsize \url{https://plantweaver.github.io}}
}

\begin{document}

\twocolumn[{%
\renewcommand\twocolumn[1][]{#1}%
\maketitle
\begin{center}
  \vspace*{-0.1em}
  \centering
  \includegraphics[width=\linewidth]{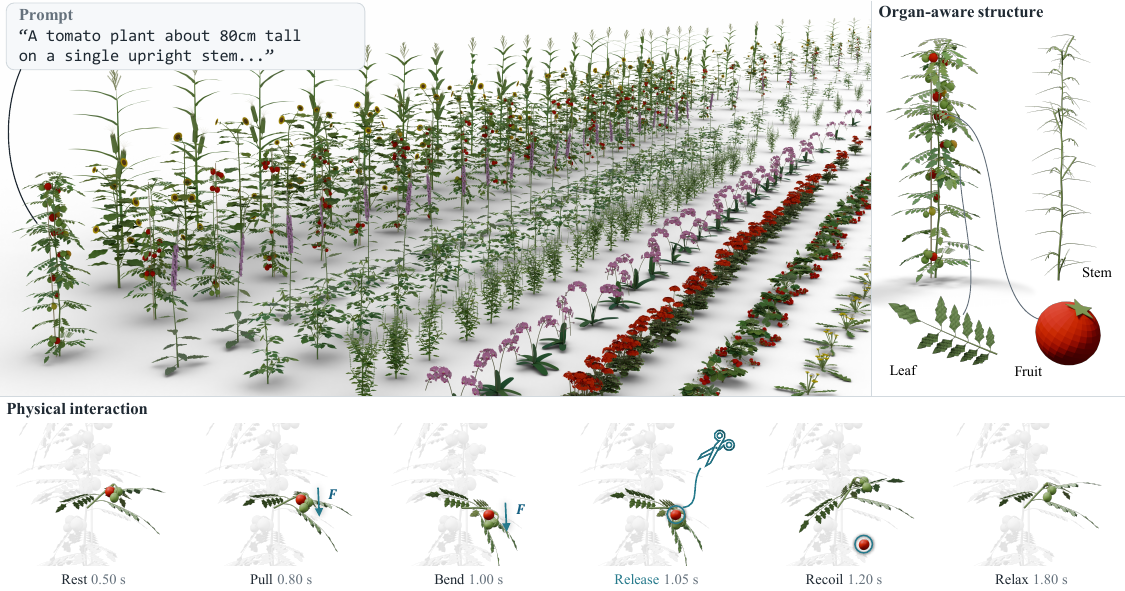}
\vspace{-1.5em}
  \captionof{figure}{\textbf{\ours} translates text prompts into reusable, botanically grounded developmental descriptions that generate diverse plant populations without repeated language-model inference (left). The generated assets preserve organ identities and connectivity (top right) and support physical simulation with flexible stems and detachable fruits (bottom), enabling the construction of diverse, structured, and simulation-ready plant populations. }
  \label{fig:teaser}
\end{center}
\vspace{1.0em}
}]

\begin{abstract}
Synthetic plant scenes support agricultural perception and robotic simulation, but authoring diverse, botanically faithful assets remains labor-intensive. We present PlantWeaver, an agentic framework for generating simulation-ready plant populations from text. An LLM agent compiles species-specific botanical profiles from web sources and reference photographs to guide autonomous authoring, validation, and revision of reusable developmental descriptions. These descriptions encode branching, organ formation, and trait distributions, allowing a shared growth engine to generate diverse individuals without further LLM inference. The plants retain organ identities and connectivity for compilation into articulated simulation assets. Experiments across species and LLMs show improved botanical correctness and comparable or better human preference than individually authored baselines, while population tests demonstrate consistency with botanical profiles. We will release our implementation and plant generator.
\end{abstract}

\section{Introduction}
\label{sec:introduction}

Synthetic plant assets support agricultural perception and robot simulation by providing controllable scenes for developing and evaluating vision systems and robots~\cite{bertoglio2025cropcraft}. Harvesting and pruning require individual organs, their attachments, and physical properties that determine their response to interaction. At the scene level, environments may contain hundreds or thousands of plants with different configurations. Creating assets for these applications involves modeling both species-specific organ structure and variation across individuals.

Large language model (LLM) agents provide a natural-language interface to 3D asset modeling through code generation, execution, and feedback~\cite{hu2024scenecraft,huang2024blenderalchemy,ll3m,sun2025threedgpt}. Simulation-aware agentic modeling extends this direction to articulated objects whose parts and joints support physical interaction~\cite{articraft}. However, extending these approaches to plants raises additional challenges: their form depends on recursive branching and species-specific arrangements of leaves, flowers, and fruit, so recognizable geometry alone does not establish correct organ structure. Procedural plant models capture this organization through developmental rules~\cite{prusinkiewicz1990abop,vos2010fspm,dereffye2021greenlab}, but tailoring these rules to each species and, where needed, its growing conditions requires botanical expertise and substantial manual effort, even in LLM-assisted workflows~\cite{hadadi2025floraforge}. Moreover, when an agent authors each individual independently, LLM inference and refinement must be repeated across the population.

We introduce \ours{}, an agentic framework that shifts LLM authoring from individual plant models to reusable developmental descriptions. Given a text prompt, an LLM agent autonomously constructs, validates, and revises a description of branching, organ formation, and trait distributions in a species-agnostic botanical vocabulary. Once finalized, a generic growth engine repeatedly executes the description with different random seeds, generating diverse individuals at scale without further LLM inference. The generated plants retain organ identities and connectivity, enabling compilation into articulated assets for physical simulation.

To support species-specific modeling, our agent harness organizes information from web sources and reference photographs into a botanical profile. The profile records organ traits, their arrangements, and the evidence supporting them, making the requirements for developmental modeling explicit before generation. The botanical profile guides description authoring and provides a shared reference for organ- and population-level validation. Together with visual review, structured validation feedback identifies discrepancies and the relevant description fields, enabling targeted revision of the rules used to generate the population.

We evaluate \ours{} across plant species and base LLMs against general-purpose coding agents, simulation-aware 3D modeling agents, and procedural generation frameworks. Despite reusing descriptions to generate diverse individuals, \ours{} achieves higher botanical correctness and comparable or better human preference than baselines that model each plant individually. Reusing descriptions reduces generation time across the population, while ablations show the benefit of search and profiling.

\para{Contributions.}
Our contributions are twofold. First, we introduce agentic authoring of reusable developmental descriptions for generating structured, simulation-ready plant populations without per-individual LLM inference. Second, we develop a botanically grounded agent harness that connects profiling from external evidence to species-specific description authoring, organ- and population-level validation, and targeted revision.

\section{Related Work}
\label{sec:related}

\para{Plant modeling and generation.}
Plant modeling encompasses rule-based generation and reconstruction from observations~\cite{okura2022plantreview}. Developmental models encode branching and organ formation through botanical rules~\cite{prusinkiewicz1990abop,boudon2012lpy,dereffye2021greenlab}, while functional--structural and biophysical frameworks connect architecture with physiological processes~\cite{vos2010fspm,bailey2019helios}. Rules and parameters are tailored to species and growing conditions, with inverse methods estimating parameters from observed geometry~\cite{stava2014inverse,zhai2025cropcraft}. Learned models represent organ and plant geometry~\cite{yang2026neuraleaf,liu2025boxplant,cheng2025demeter}, while diffusion-based pipelines refine geometry or synthesize views for reconstruction and generation~\cite{hartley2025plantdreamer,hartley2026plantorbiter}. Building on developmental modeling, we focus on reusable descriptions that generate organ-structured populations without repeated LLM inference.

\para{LLM-assisted procedural modeling.}
LLMs control generators and write modeling code~\cite{sun2025threedgpt,hu2024scenecraft,huang2024blenderalchemy,ll3m}, with procedural systems and compositional interfaces providing reusable building blocks~\cite{raistrick2023infinigen,raistrick2026procfunc}. Benchmarks assess geometric and structural correctness and tool use~\cite{threedcodebench,threedharnessbench,p3dbench}. For plants, FloraForge combines expert-guided LLM co-design with reusable generators and editable organ parameters~\cite{hadadi2025floraforge}. Vision-based methods infer simulator configurations, organ architecture, or parameters for existing generators~\cite{vlmhelios,cowpea,hadadi2026vlmmaize}. Our contribution is autonomous developmental authoring grounded in botanical profiles from external evidence, with organ-level and population-level checks and visual feedback guiding targeted revision.

\para{Simulation-ready plant modeling and interaction.}
Articulated assets are generated procedurally~\cite{joshi2025infinigenarticulated} or reconstructed from observations~\cite{chen2024urdformer,real2code,urdfanything}, while LLM-based systems automate geometry, articulation, and physical authoring~\cite{le2025articulate,lam,articraft,shui2026articad,zhang2026usdcraft}. Simulator feedback and mechanism graphs support physical and functional validation~\cite{cai2026artisan,yang2026marc}. Plants require distributed compliance and attachment of many organs. Agricultural simulation research addresses scene generation~\cite{bertoglio2025cropcraft}, compliant plant modeling and dynamics~\cite{barthelme2023dynamicplants,yao2024femplant,jacob2024branchdynamics}, fruit detachment~\cite{deng2024gazeboplants}, and deformation-aware manipulation~\cite{kim2024treemanipulation,yao2025safeleaf,subedi2026findfruit}. We connect autonomous developmental authoring to physical compilation using plant biomechanics~\cite{niklas1992biomechanics}, enabling scalable generation of structured, simulation-ready plant populations.

\begin{figure*}[t]
    \centering
    \includegraphics[width=\linewidth]{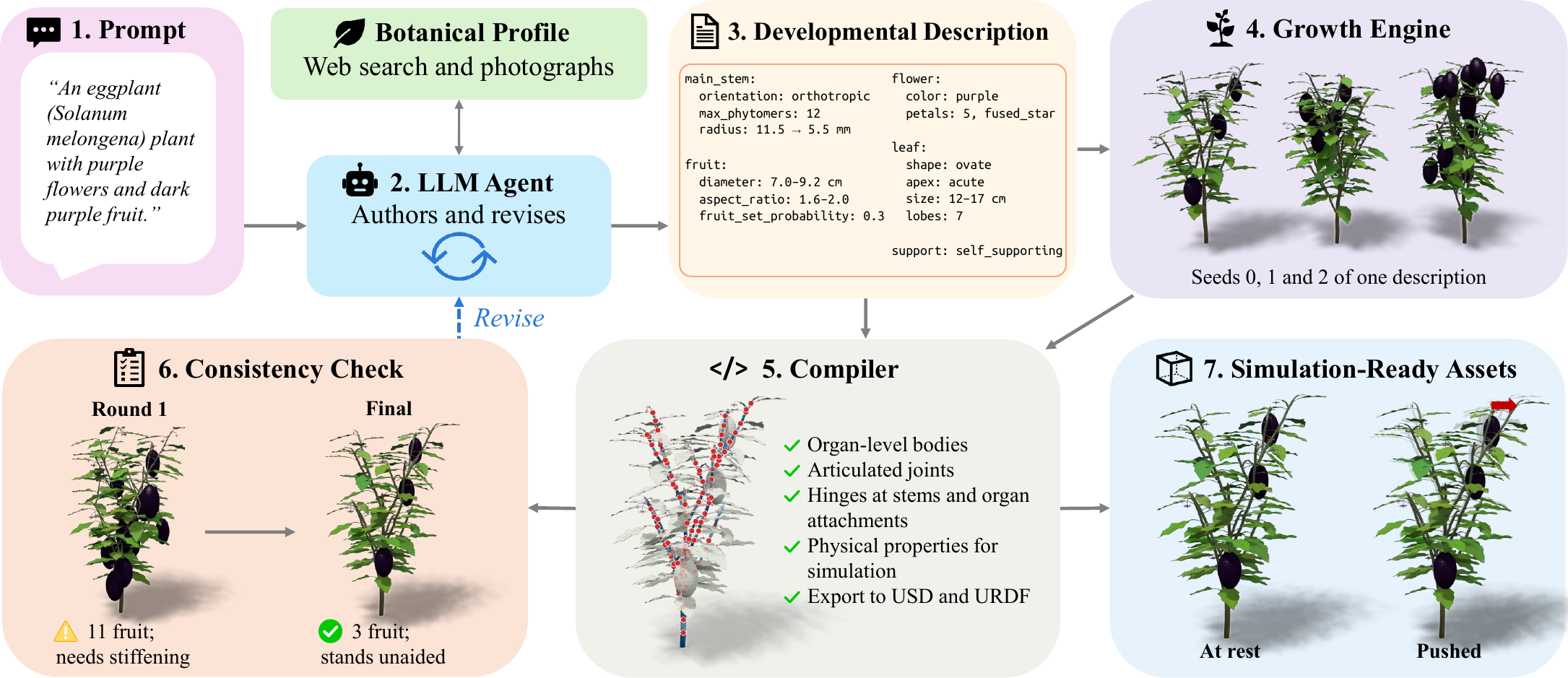}
    \caption{
    \textbf{Overview of \ours.}
An LLM agent compiles a botanical profile from web sources and reference photographs, then authors a reusable developmental description in a species-agnostic botanical vocabulary (excerpt shown). Organ- and population-level validation and visual feedback guide targeted revision. Once finalized, the description supports the generation of diverse, simulation-ready individuals through a shared growth engine and compiler, without further LLM inference.
    }
    \label{fig:pipeline}
\end{figure*}

\section{Method: \ours}
\label{sec:method}

\para{Formulation.}
Given a text prompt $q$ describing a plant, \ours constructs a reusable developmental description $D$ from which simulation-ready individuals are generated (\cref{fig:pipeline}). The description specifies branching, organ formation, and trait distributions in a species-agnostic botanical vocabulary. For each individual $i$, the growth engine $\mathcal{G}$ executes the description with a random seed $s_i$ to produce a growth trace $T_i$, which records its organ structure, connectivity, and spatial arrangement. A compiler $\mathcal{C}$ constructs the corresponding simulation asset $A_i$:
\begin{equation}
    T_i = \mathcal{G}(D,s_i), \quad
    A_i = \mathcal{C}(T_i,D).
\end{equation}
Each asset $A_i$ contains organ geometry and identities, rigid bodies, joints, and physical properties supporting stem bending and fruit detachment. Once the description $D$ is fixed, additional individuals are generated by changing the seed without further LLM inference.

\para{Overview.}
An LLM agent autonomously authors, validates, and revises the developmental description $D$. It first compiles a botanical profile from web sources and reference photographs, making the requested plant's organ traits and arrangements explicit. It then authors the description and uses the growth engine $\mathcal{G}$ and compiler $\mathcal{C}$ to generate individuals for validation. Organ- and population-level validation and visual review identify discrepancies with the profile and prompt, guiding targeted revisions.

\subsection{Developmental plant generation}
\label{sec:generation}

\para{Developmental description.}
Following rule-based and developmental plant models~\cite{prusinkiewicz1990abop,dereffye2021greenlab}, the developmental description $D$ represents plant development through shoot archetypes. Each archetype defines a type of shoot, such as a main stem, lateral branch, or flowering branch, as a sequence of phytomers: repeated units comprising an internode and a node bearing leaves and axillary buds. An initial archetype starts growth, while references from buds to archetypes determine how new shoots branch. The growth engine $\mathcal{G}$ and compiler $\mathcal{C}$ interpret this shared vocabulary without species-specific code.

Each archetype specifies growth and posture, phyllotaxis (leaf arrangement), meristem fates, organ morphology, and physical traits. Morphological traits describe leaves, inflorescences, flowers, and fruit. Physical traits include material moduli, twisting-to-bending compliance ratios, fruit density, and detachment force. Values may be fixed or sampled from specified distributions, encoding variation in dimensions, branching, and organ counts. Plant-level targets specify counts and heights, while support conditions describe how the plant is supported. The full specification and a complete example of the developmental description $D$ are provided in the supplementary material. %

\para{Population generation.}
The growth engine $\mathcal{G}$ produces a growth trace $T_i$ by executing the shoot archetypes in the description $D$ with seed $s_i$. At each shoot tip, the apical meristem resolves to a vegetative, reproductive, or dormant state, controlling shoot extension and termination. Axillary buds produce shoots of their referenced archetypes, while reproductive buds and apices form inflorescences. Positions within an inflorescence are assigned to fruit, open flowers, or buds. These developmental choices and organ traits are sampled using the seed, so the same description and seed reproduce the same trace, while different seeds generate variation among individuals.

The targets stored in the description $D$ constrain this process. Explicit organ counts are allocated among admissible buds and reproductive positions; requests that cannot be realized are reported through validation feedback. For range-valued targets, population-level scaling brings typical counts and heights into the requested ranges while preserving individual variation. The realized organ structure, attributes, and positions remain in the growth trace $T_i$ for compilation and botanical validation.

\para{Simulation-ready compilation.}
The compiler $\mathcal{C}$ converts the growth trace $T_i$ into a simulation asset $A_i$ using the morphological and physical traits specified in the description $D$. It constructs organ geometry while preserving the identities and connectivity recorded in the trace. Stems are meshed as tapered tubes whose cross-sections follow the supported leaf area through the pipe model~\cite{shinozaki1964pipe}. Leaves, flowers, inflorescence stalks, and fruit are meshed from their traits and placed according to the trace. Rigid internode links are connected by joints allowing bending and twisting; other organs and their stalks are attached through their own joints.

Joint stiffnesses follow beam-based relations using organ dimensions and the material traits in the description $D$~\cite{niklas1992biomechanics,vogel1992twist}, while mesh volumes and densities determine masses and inertias. Spring preloads balance gravity at the pose recorded in the growth trace $T_i$; stability correction or settling then follows the support conditions in the description. Fruit joints use the specified force as their detachment threshold. The resulting simulation asset $A_i$ is exported as an OpenUSD physics scene~\cite{openusd} for use in Isaac Sim and MuJoCo~\cite{todorov2012mujoco}.

\subsection{Botanically grounded agent harness}
\label{sec:harness}

\para{Botanical profiling.}
\label{sec:profile}
The agent first compiles a profile for the plant and developmental stage requested in the prompt $q$. It records growth habit, branching, phyllotaxis, and organ morphology, together with ranges for plant dimensions, organ sizes, and physical traits. This profile is the shared reference for authoring the developmental description $D$ and validating its generated individuals.

The agent gathers information from web sources and reference photographs. Each profile entry records its basis: the prompt, a cited source, a photograph, the LLM's knowledge, or inference from a named relative. The agent uses explicit source statements to set the corresponding traits, while photographs help resolve visual details left unspecified by text. The harness flags internal contradictions, such as an unbranched habit accompanied by side shoots, for the agent to correct. The agent then authors the description $D$ using the profile and developmental schema.

\para{Organ- and population-level validation.}
\label{sec:checks}
After each update to the description $D$, the harness runs the growth engine $\mathcal{G}$ to produce a growth trace $T_i$ and the compiler $\mathcal{C}$ to construct a simulation asset $A_i$. Programmatic checks compare organ traits in the trace with the profile and assess prompt targets, geometric consistency, and physical behavior. For each trait mismatch, validation feedback reports the expected state, the realized state, and the description fields to revise.

To validate the description $D$ across a population of individuals, the harness uses the growth engine $\mathcal{G}$ to sample additional growth traces $T_i$ with different seeds and compares their traits with the same profile. These population-level checks do not invoke the compiler $\mathcal{C}$, so geometry and physics assets are not constructed. A description is flagged when more than a prescribed fraction of sampled individuals disagree with the profile. This identifies descriptions whose reviewed individual satisfies the requirements but whose other samples do not.

\para{Visual review and targeted revision.}
The agent receives validation results and measurements alongside whole-plant and organ-level renderings of the simulation asset $A_i$ and the reference photographs. It reviews growth habit, stems, leaves, inflorescences, flowers, fruit, and overall size, identifying discrepancies, supporting evidence, and the relevant fields of the developmental description $D$. The agent then accepts the description or edits selected fields, archetypes, targets, or support conditions, after which the harness repeats generation and validation. Features beyond the vocabulary's scope are reported explicitly. If the profile itself misrepresents the requested plant, the agent may correct it while recording the reason for the change. Detailed tool interfaces, review instructions, and validation settings are provided in the supplementary material.

\section{Experiments}
\label{sec:exp}

We evaluate \ours in terms of botanical correctness, prompt match, and human preference. We compare against \emph{mass-grown} systems that reuse generator specifications and \emph{tailor-made} systems that author each plant individually. We further examine the contribution of botanical grounding through ablations and evaluate the consistency and efficiency of population generation from one description. Physical simulation evaluations are provided in the supplementary material.

\subsection{Experimental setup}
\label{sec:exp:setup}

\para{Plant species.}
We evaluate on 22 plant species: 16 \emph{well-documented} species (\eg,~avocado) and six \emph{less-documented} species (\eg, \emph{Asteropyrum peltatum}) with limited online botanical information. We compile reference traits for quantitative evaluation from the Flora of China~\cite{floraofchina} and the Flora of North America~\cite{floraofnorthamerica}. 

For each species, we prepare a \emph{species} prompt naming the species and developmental stage, and a \emph{detailed} prompt additionally specifying height and organ counts, as illustrated below. Species prompts assess botanical correctness from minimal input, while detailed prompts assess prompt match over 66 numerical constraints across the 22 species. Each quantitative prompt is run once per method and base LLM. Complete prompts are provided in the supplementary material. %

\smallskip
\noindent\fbox{%
\begin{minipage}{\dimexpr\linewidth-2\fboxsep-2\fboxrule\relax}
\small
\textbf{Species prompt}\par
A young avocado (\emph{Persea americana}) tree with fruit

\medskip
\textbf{Detailed prompt}\par
An avocado (\emph{Persea americana}) tree about 2.5 m tall with 3 main branches and 6 fruit
\end{minipage}%
}
\smallskip

\vspace{-3mm}
\para{Baselines.}
We compare two categories of baselines:

\emph{Mass-grown} systems reuse generator specifications across individuals like ours. The baseline in this category is FloraForge~\cite{hadadi2025floraforge}, for which the LLM sets the parameters of released plant templates with the generator source provided as context.

\emph{Tailor-made} systems author each plant through a separate LLM run. We compare with Articraft~\cite{articraft} for simulation-ready agentic modeling, 3D-GPT~\cite{sun2025threedgpt} for LLM-assisted procedural modeling, and generic LLM coding agents using Codex and DeepSeek Harness. The generic agents are asked to produce OpenUSD plant assets with explicit organs and articulation, using a shell and web search. Baselines follow their released workflows, with implementation details and model-provider adaptations provided in the supplementary material. %

\para{Base LLMs.}
We use GPT-6-astra and GPT-5.5 through OpenAI's API, and self-host the open-weight Qwen3.8-27B. Comparisons are conducted within each base LLM, including for the Codex and DeepSeek Harness baselines, to separate the effect of the modeling framework from that of the underlying model. Botanical correctness and prompt match are evaluated across these three base LLMs; 3D-GPT is evaluated only with GPT-6-astra. Detailed inference settings are provided in the supplementary material. %

\para{Evaluation metrics.} We assess generated plants using three metrics:

\emph{Botanical correctness} counts agreement with 182 species-specific reference traits compiled from the Flora of China~\cite{floraofchina} and the Flora of North America~\cite{floraofnorthamerica} across the 22 species. These traits cover branching, leaf morphology and arrangement, and inflorescence structure; the complete list is provided in the supplementary material. We score traits of organs present in the reference plants, excluding absent-organ traits; failed generations receive zero.

\emph{Prompt match} counts satisfied height and organ-count constraints, totaling 66 across the detailed prompts. Exact numerical requests require exact matches, while ``about'' allows a tolerance of $\pm20\%$. For both metrics, the same rule-based organ reader extracts traits and measurements from the exported OpenUSD assets of all methods. Reference construction and scoring details are provided in the supplementary material.

\emph{Human preference} reports pairwise preference rates for \ours against each baseline and Bradley--Terry (BT) scores fitted across all method pairs. Preference rates exclude ties; 50\% indicates equal preference. For BT fitting, a tie contributes half a win to each method, and the estimated strengths are normalized to sum to one. We also report generation success as the number of prompts yielding an asset.

\para{Human evaluation.}
\label{sec:exp:survey}
To complement trait-based quantitative evaluation with judgments of whole-plant and organ appearance, we use a separate set of 24 crop and garden species. All systems use GPT-6-astra as the base LLM. For each prompt, we evaluate all pairs of methods that produce an asset, using one generated individual per method and randomized presentation order. For each pair, raters inspect whole-plant and organ-level comparisons, then provide a single overall preference judgment. We recruit 10 raters and collect 10 judgments per pair. These judgments support both direct preference comparisons and BT scoring across methods. Detailed generation settings and evaluation protocols are provided in the supplementary material.

\subsection{Main results}
\label{sec:exp:compare}

\begin{table*}[t]
  \centering
  \caption{\textbf{Quantitative comparison across base LLMs.} Best scores within each base LLM are in bold (ties included). Generated and botanical correctness use species prompts for 22 species; prompt match uses detailed prompts and is evaluated across the same three base LLMs. Header parentheses give the number of prompts (generated), reference traits (botanical correctness), or numerical constraints (prompt match).}\vspace{-3mm}
  \label{tab:astra}
  \label{tab:compare}
  \label{tab:specified}
  \footnotesize
  \setlength{\tabcolsep}{3.5pt}
  \begin{NiceTabular*}{\linewidth}{@{\hspace{4pt}\extracolsep{\fill}}lllccccc@{\hspace{4pt}}}[cell-space-limits=1pt]
    \CodeBefore
      \rowcolor{black!4}{1,2}
      \tikz \fill [cyan!6] ([yshift=0.5pt]8-|3) rectangle ([yshift=3.5pt]9-|9) ;
      \tikz \fill [cyan!6] ([yshift=0.5pt]13-|3) rectangle ([yshift=3.5pt]14-|9) ;
      \tikz \fill [cyan!6] ([yshift=0.5pt]18-|3) rectangle ([yshift=3.5pt]19-|9) ;
    \Body
    \toprule
    & & & & \Block{1-3}{Botanical correctness} & & & Prompt match \\
    \cmidrule(lr){5-7}
    Base LLM & Category & Method & \shortstack{Generated\\(22)}
      & \shortstack{Well-documented\\(132)}
      & \shortstack{Less-documented\\(50)}
      & \shortstack{All\\(182)}
      & \shortstack{Constraints met\\(66)} \\
    \midrule
    \Block[l]{6-1}{GPT-6-astra} & \Block[l]{4-1}{Tailor-made} & Codex & \textbf{22} & 105 & 37 & 142 & 58 \\
     &  & DeepSeek Harness & \textbf{22} & 101 & 37 & 138 & 57 \\
     &  & Articraft~\cite{articraft} & \textbf{22} & 6 & 7 & 13 & 24 \\
     &  & 3D-GPT~\cite{sun2025threedgpt} & 0 & 0 & 0 & 0 & 0 \\
    \cmidrule(lr){2-8}
     & \Block[l]{2-1}{Mass-grown} & FloraForge~\cite{hadadi2025floraforge} & 1 & 1 & 0 & 1 & 3 \\
     &  & \ours & \textbf{22} & \textbf{116} & \textbf{41} & \textbf{157} & \textbf{59} \\
    \midrule
    \Block[l]{5-1}{GPT-5.5} & \Block[l]{3-1}{Tailor-made} & Codex & \textbf{22} & 81 & 28 & 109 & 54 \\
     &  & DeepSeek Harness & \textbf{22} & 82 & 30 & 112 & 54 \\
     &  & Articraft~\cite{articraft} & \textbf{22} & 7 & 7 & 14 & 24 \\
    \cmidrule(lr){2-8}
     & \Block[l]{2-1}{Mass-grown} & FloraForge~\cite{hadadi2025floraforge} & 4 & 1 & 0 & 1 & 8 \\
     &  & \ours & \textbf{22} & \textbf{114} & \textbf{37} & \textbf{151} & \textbf{55} \\
    \midrule
    \Block[l]{5-1}{Qwen3.8-27B} & \Block[l]{3-1}{Tailor-made} & Codex & \textbf{21} & 41 & 19 & 60 & 44 \\
     &  & DeepSeek Harness & 4 & 9 & 7 & 16 & 8 \\
     &  & Articraft~\cite{articraft} & 18 & 4 & 5 & 9 & 20 \\
    \cmidrule(lr){2-8}
     & \Block[l]{2-1}{Mass-grown} & FloraForge~\cite{hadadi2025floraforge} & 17 & 3 & 7 & 10 & 17 \\
     &  & \ours & \textbf{21} & \textbf{102} & \textbf{27} & \textbf{129} & \textbf{55} \\
    \bottomrule
  \end{NiceTabular*}\vspace{-3mm}
\end{table*}

\begin{table}[t]
  \centering
  \caption{\textbf{Human preference.} 24-species study with 10 human raters. \emph{Ours preferred}: percentage of decisive judgments (ties excluded) favoring \ours over each baseline. \emph{BT score}: Bradley--Terry strength fitted on the judgments of all method pairs, with a tie counted as half a win for each method, normalized to sum to one.}\vspace{-3mm}
  \label{tab:human}
  \footnotesize
  \setlength{\tabcolsep}{4pt}
  \begin{NiceTabular*}{\linewidth}{@{\hspace{4pt}\extracolsep{\fill}}llcc@{\hspace{4pt}}}[cell-space-limits=1.5pt]
    \CodeBefore
      \rowcolor{black!4}{1}
      \tikz \fill [cyan!6] ([yshift=0.5pt]6-|2) rectangle ([yshift=3.5pt]7-|5) ;
    \Body
    \toprule
    Category & Method & Ours preferred (\%) & BT score \\
    \midrule
    \Block[l]{3-1}{Tailor-made} & Codex & 60.9 & 0.182 \\
     & DeepSeek Harness & 59.1 & 0.204 \\
     & Articraft & 52.8 & 0.284 \\
    \midrule
    \Block[l]{2-1}{Mass-grown} & FloraForge & 63.2 & 0.043 \\
     & \ours & -- & \textbf{0.287} \\
    \bottomrule
  \end{NiceTabular*}
\end{table}

\para{Quantitative and visual comparisons.}
\ours leads on every quantitative metric with both frontier and open-weight base LLMs, achieving the highest botanical correctness and prompt match while matching the best generation success (\cref{tab:astra}). Against the mass-grown baseline FloraForge, botanical correctness rises from 1 to 157 matched traits with GPT-6-astra and from 10 to 129 with Qwen3.8-27B, and more prompts yield an asset. \ours also outperforms tailor-made baselines that devote a separate LLM run to each individual. It matches 157/182 traits versus 142 for the strongest such baseline, Codex, with GPT-6-astra; this advantage widens with Qwen3.8-27B (129 versus 60). On detailed prompts, \ours satisfies 59/66 numerical constraints with GPT-6-astra and 55/66 with GPT-5.5, exceeding the strongest baselines at 58 and 54, respectively. The advantage in prompt match also holds with Qwen3.8-27B (55/66 versus 44 for Codex).

\begin{figure*}[t!]
\vspace{5mm}
  \centering
  \footnotesize
  \setlength{\tabcolsep}{1.5pt}
  \newlength{\PWComparisonLLMWidth}
  \setlength{\PWComparisonLLMWidth}{0.12\linewidth}
  \newlength{\PWComparisonWidth}
  \setlength{\PWComparisonWidth}{\dimexpr(\linewidth-\PWComparisonLLMWidth-12\tabcolsep)/5\relax}
  \newlength{\PWComparisonHeight}
  \setlength{\PWComparisonHeight}{\dimexpr(\textheight-60mm)/6\relax}
  \newlength{\PWComparisonSide}
  \ifdim\PWComparisonWidth<\PWComparisonHeight
    \setlength{\PWComparisonSide}{\PWComparisonWidth}
  \else
    \setlength{\PWComparisonSide}{\PWComparisonHeight}
  \fi
  \newcommand{\PWComparisonBlank}[1]{%
    \parbox[c][\PWComparisonSide][c]{\PWComparisonSide}{%
      \centering\scriptsize#1%
    }%
  }
  \gdef\PWInsetY{0.6}
  \newcommand{\PWComparisonImage}[2][\textcolor{red}{\textit{TBD}}]{%
    \raisebox{\dimexpr\depth/2-\height/2\relax}{%
      \makebox[\PWComparisonWidth][c]{%
        \IfFileExists{figs/main_comparison/#2.jpg}{%
          \includegraphics[width=\PWComparisonSide,height=\PWComparisonSide]{figs/main_comparison/#2.jpg}%
          \IfFileExists{figs/main_comparison/#2_inset.jpg}{%
            \llap{%
              \raisebox{\PWInsetY\PWComparisonSide}{%
                \setlength{\fboxsep}{0pt}\setlength{\fboxrule}{0.4pt}%
                \fcolorbox{black!40}{white}{\includegraphics[width=\dimexpr0.4\PWComparisonSide-0.8pt\relax]{figs/main_comparison/#2_inset.jpg}}%
              }%
              \hspace{0.6\PWComparisonSide}%
            }%
          }{}%
        }{%
          \PWComparisonBlank{#1}%
        }%
      }%
    }%
  }
  \newcommand{\PWComparisonPrompt}[1]{%
    \multicolumn{6}{l}{\strut\gdef\PWInsetY{0.6}\textbf{Prompt:} #1}\\[1pt]%
  }
  \newcommand{\PWComparisonPromptLow}[1]{%
    \multicolumn{6}{l}{\strut\gdef\PWInsetY{0}\textbf{Prompt:} #1}\\[1pt]%
  }
  \begin{NiceTabular}{p{\PWComparisonLLMWidth}*{5}{c}}[cell-space-limits=0pt]
    \toprule
     & \Block{1-3}{Tailor-made} & & & \Block{1-2}{Mass-grown} & \\
    \cmidrule(lr){2-4}\cmidrule(lr){5-6}
    Base LLM & Codex & DeepSeek Harness & Articraft & FloraForge & \ours \\
    \midrule
    \PWComparisonPromptLow{A flowering rosemary (\emph{Salvia rosmarinus}) shrub}
    GPT-6-astra & \PWComparisonImage{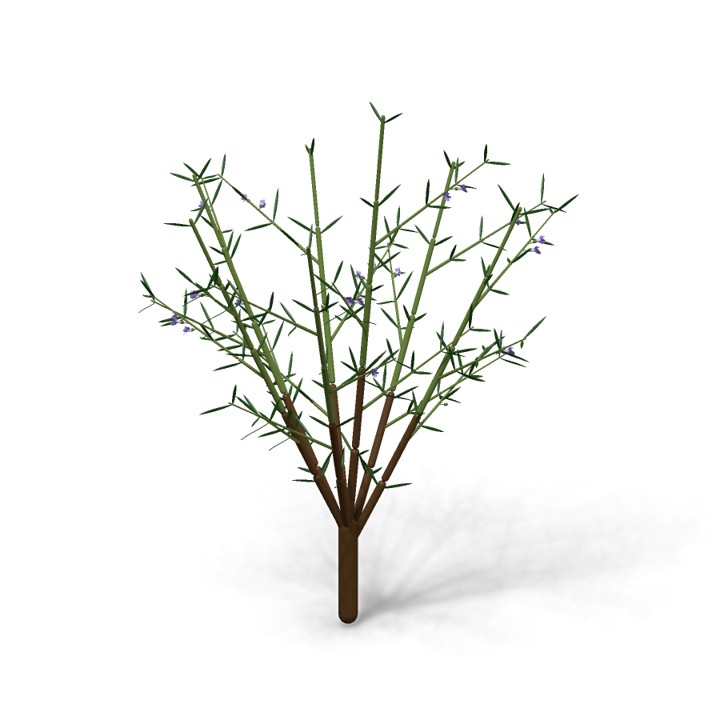} & \PWComparisonImage{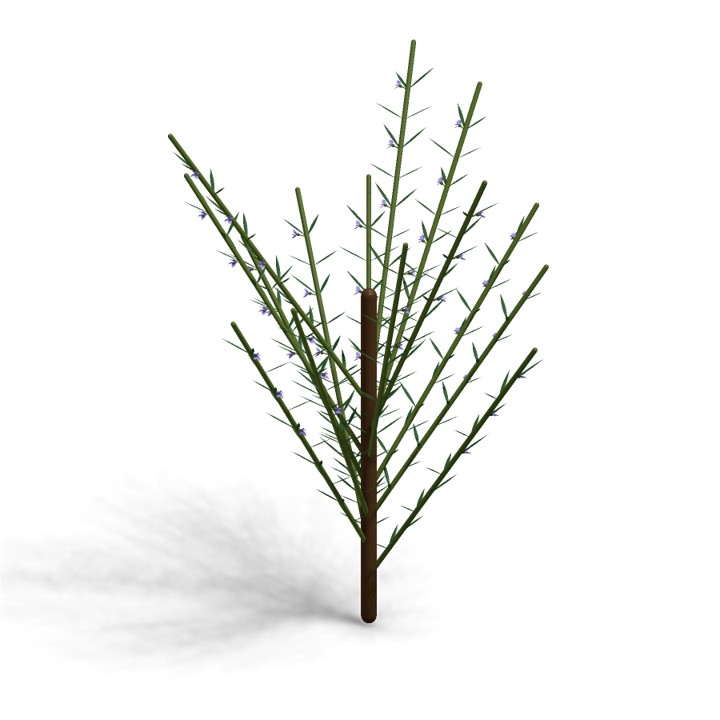} & \PWComparisonImage{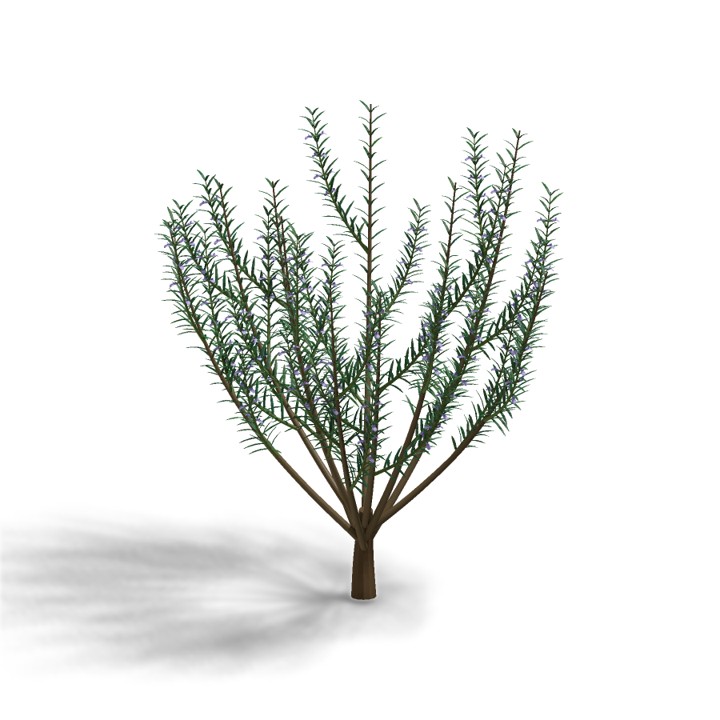} & \PWComparisonImage[Generation failed]{gpt6_astra/rosemary_floraforge} & \PWComparisonImage{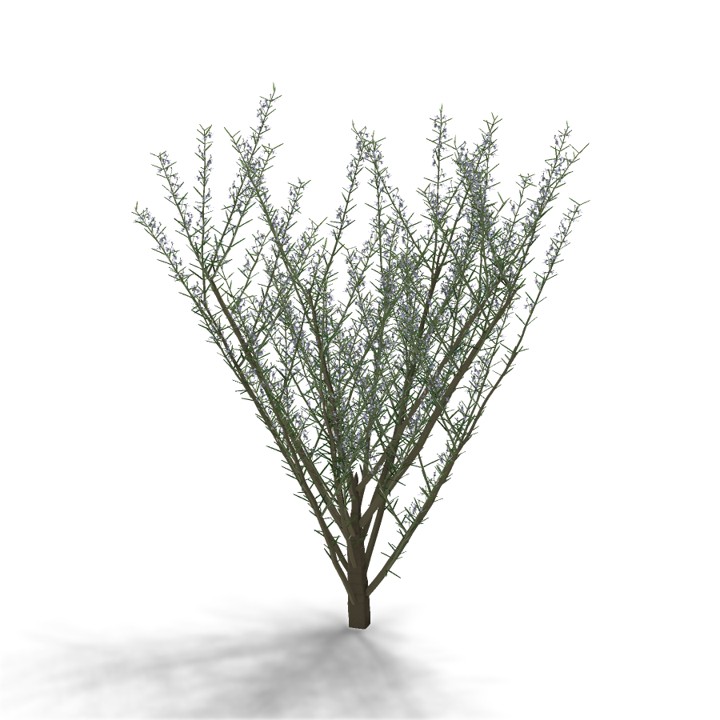} \\[2pt]
    Qwen3.8-27B & \PWComparisonImage{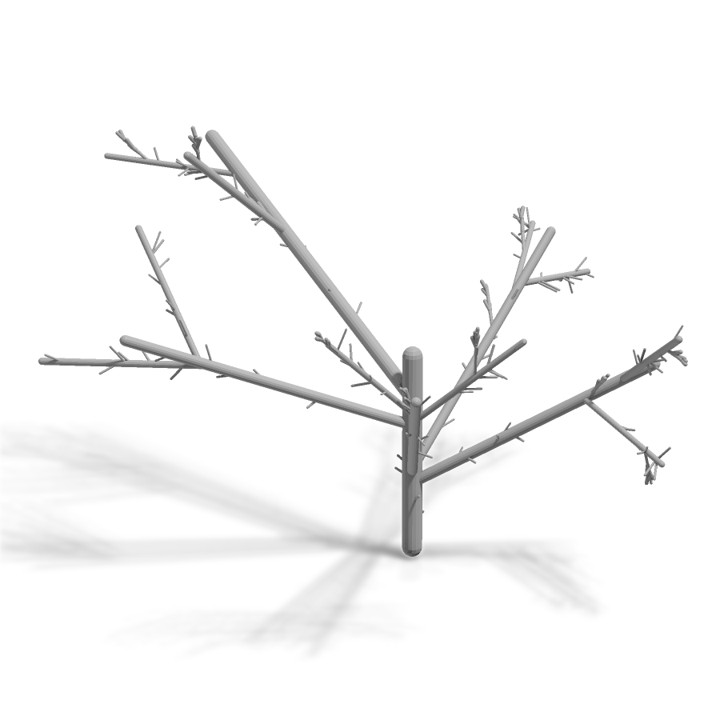} & \PWComparisonImage[Generation failed]{qwen38_27b/rosemary_deepseek} & \PWComparisonImage{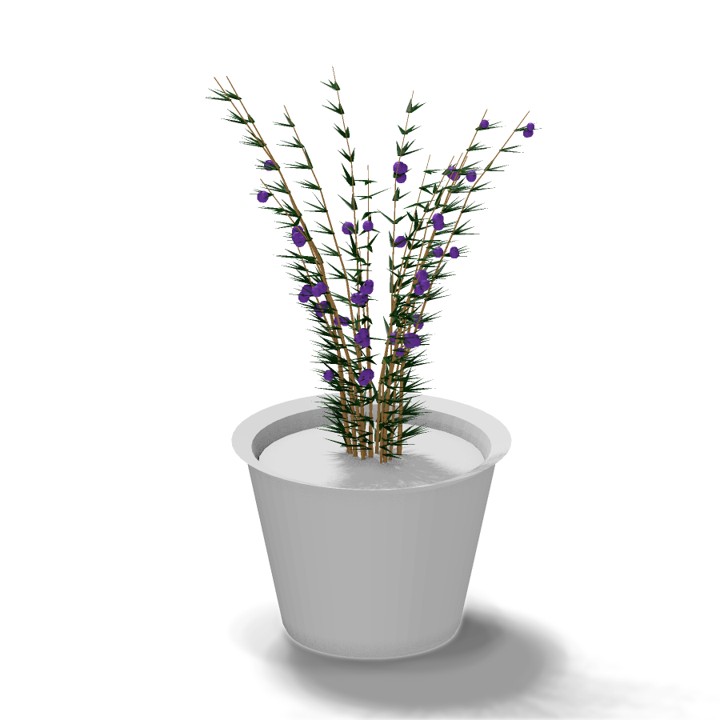} & \PWComparisonImage{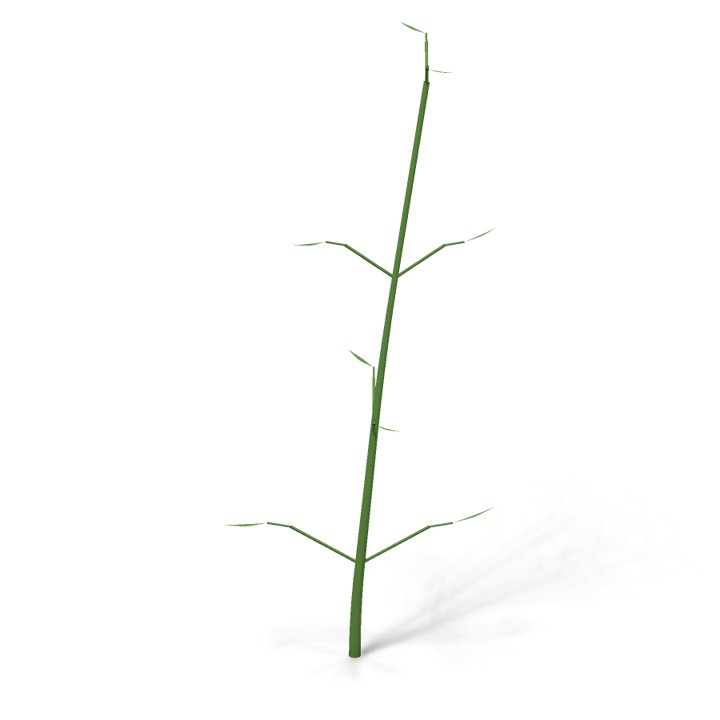} & \PWComparisonImage{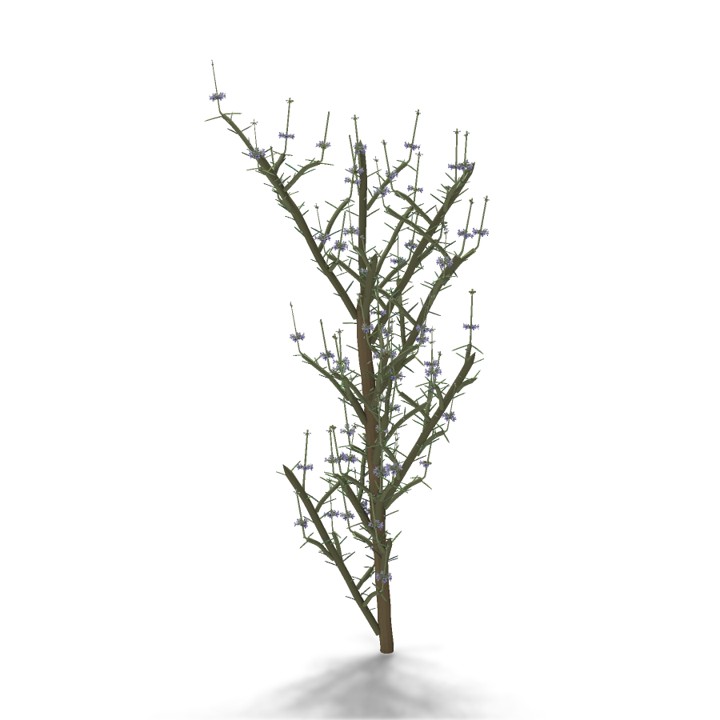} \\
    \midrule
    \PWComparisonPrompt{A flowering \emph{Beesia calthifolia} plant about 40 cm tall with 3 basal leaves and one flowering stem}
    GPT-6-astra & \PWComparisonImage{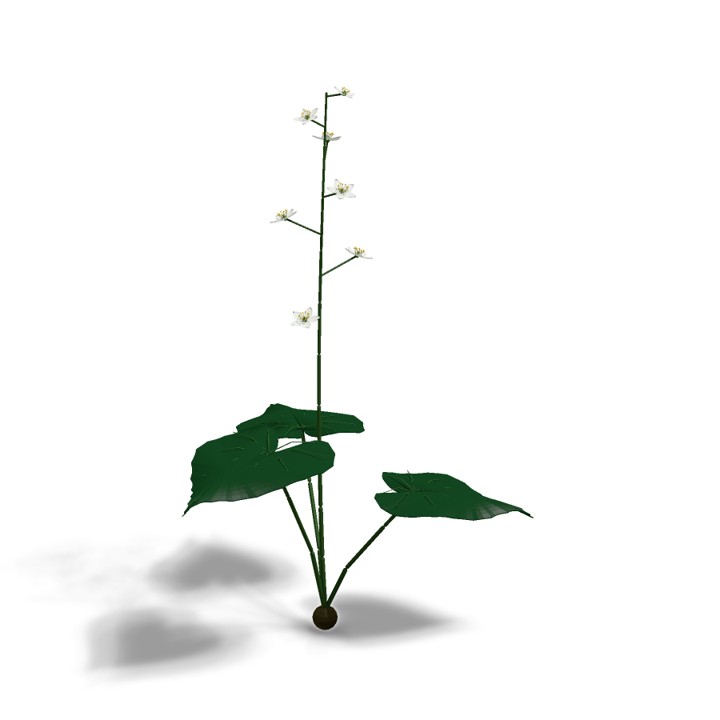} & \PWComparisonImage{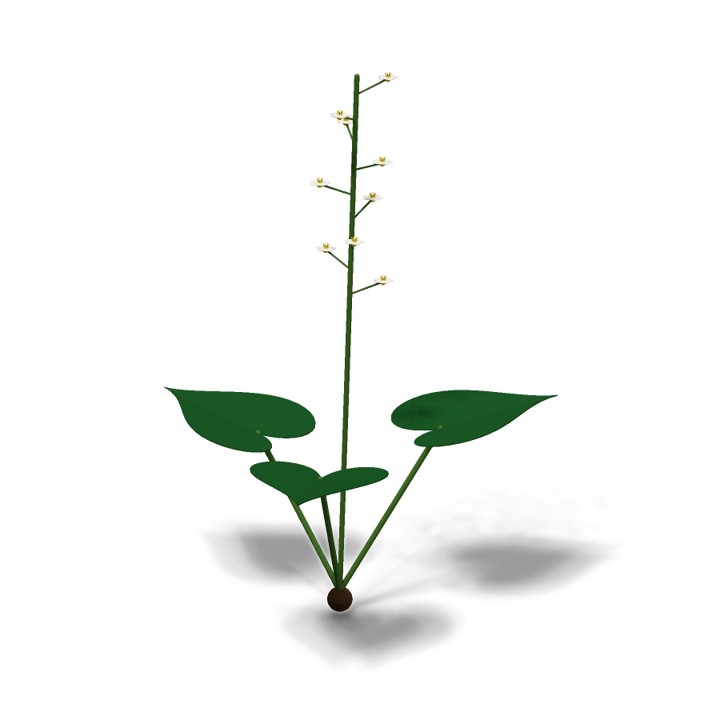} & \PWComparisonImage{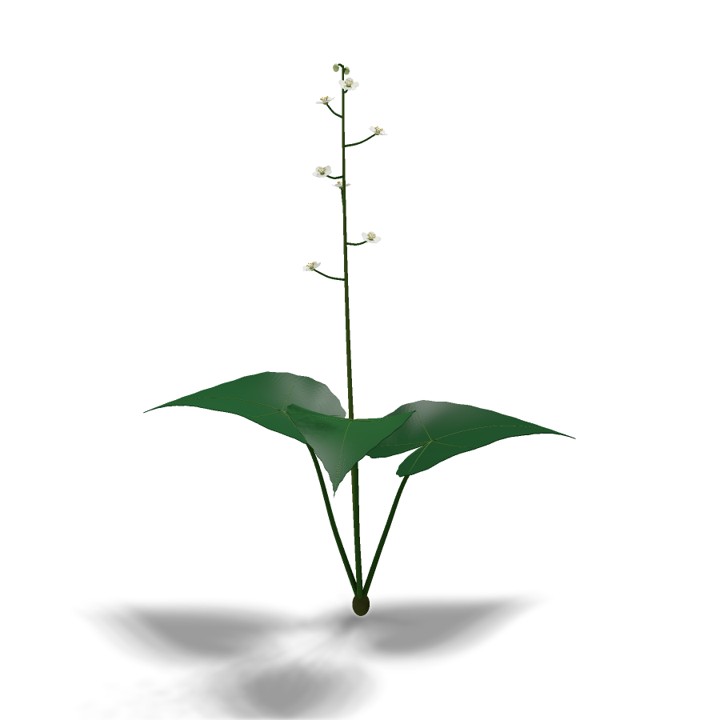} & \PWComparisonImage[Generation failed]{gpt6_astra/beesia_detailed_floraforge} & \PWComparisonImage{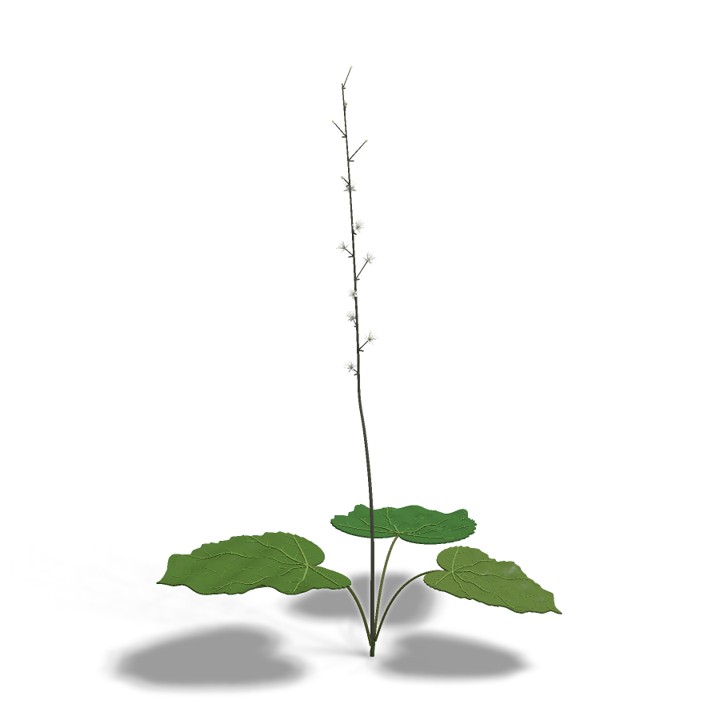} \\[2pt]
    Qwen3.8-27B & \PWComparisonImage[Generation failed]{qwen38_27b/beesia_detailed_codex} & \PWComparisonImage{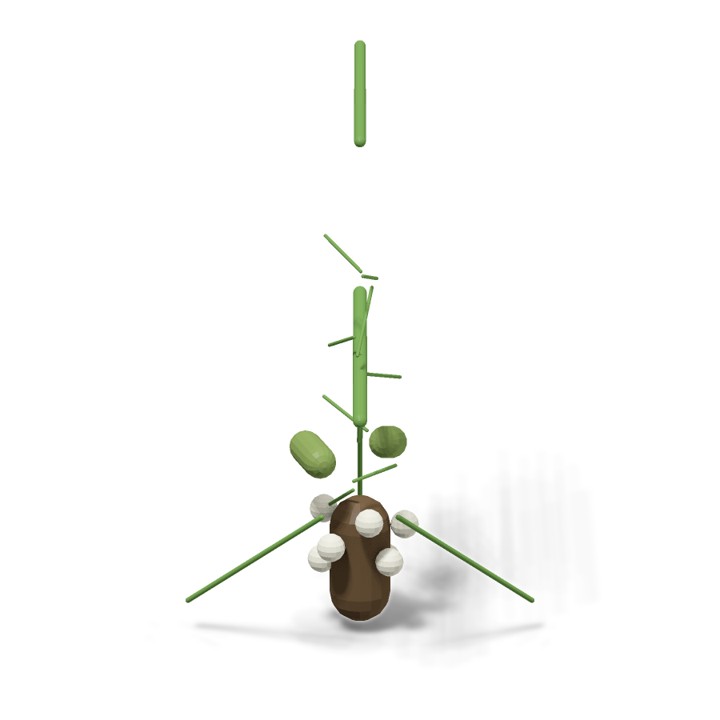} & \PWComparisonImage{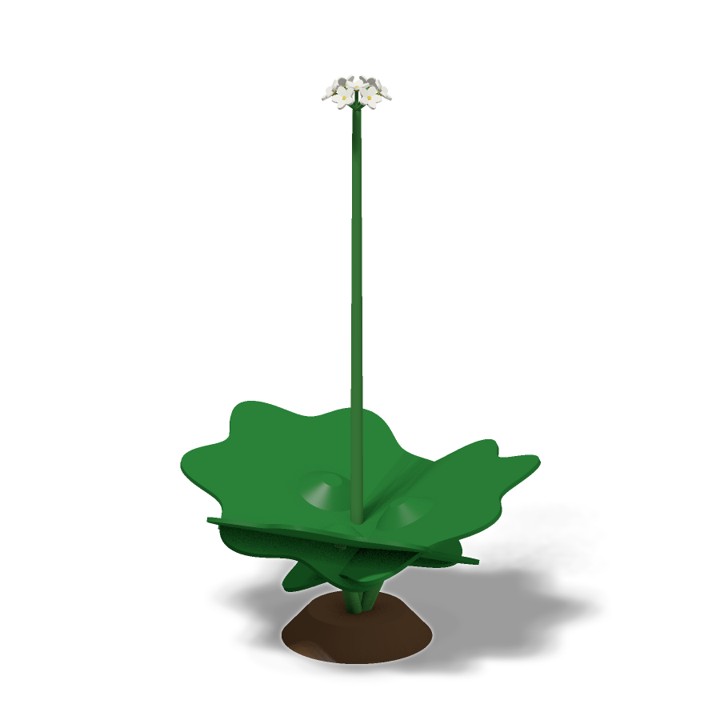} & \PWComparisonImage{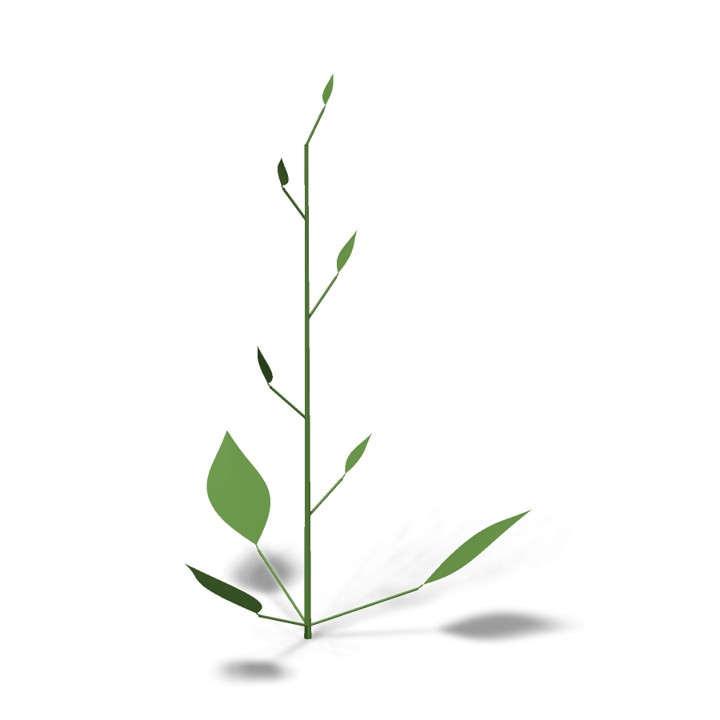} & \PWComparisonImage{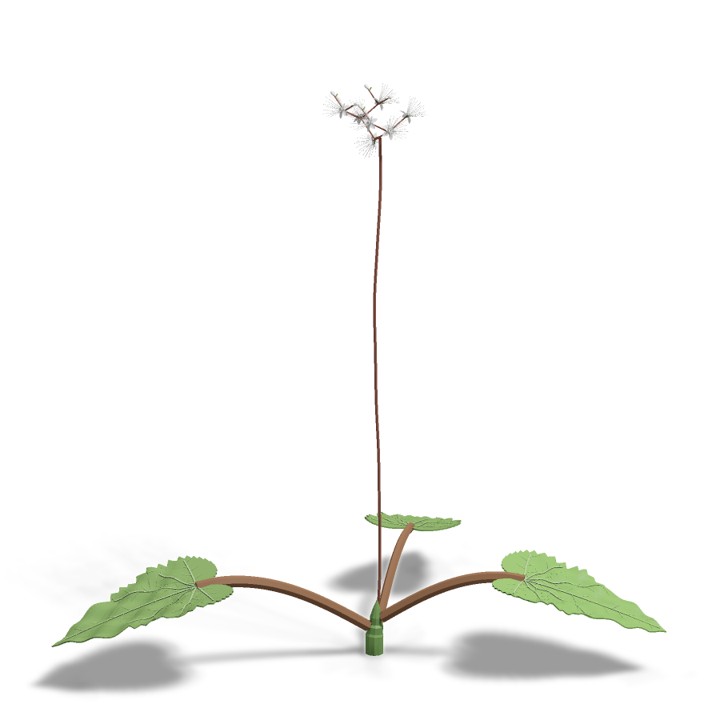} \\
    \midrule
    \PWComparisonPrompt{A flowering \emph{Mukdenia rossii} plant}
    GPT-6-astra & \PWComparisonImage{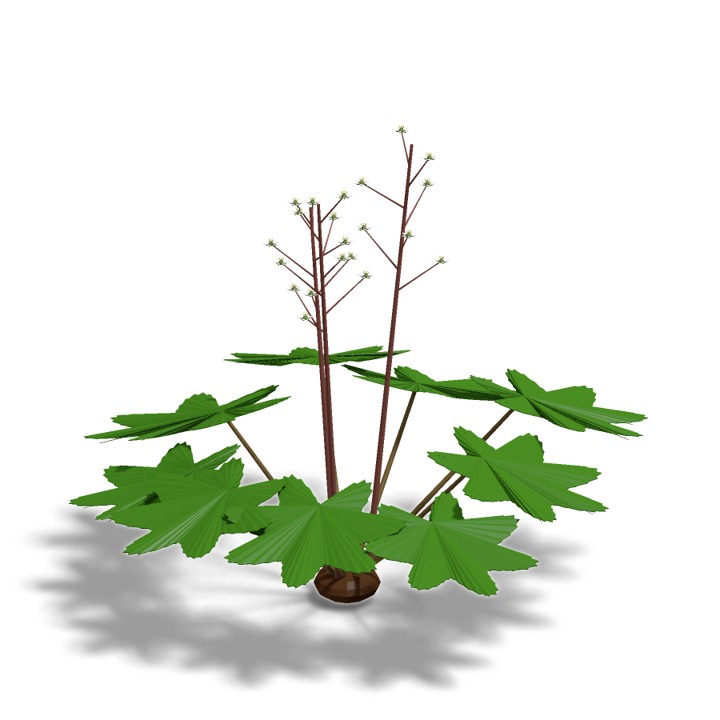} & \PWComparisonImage{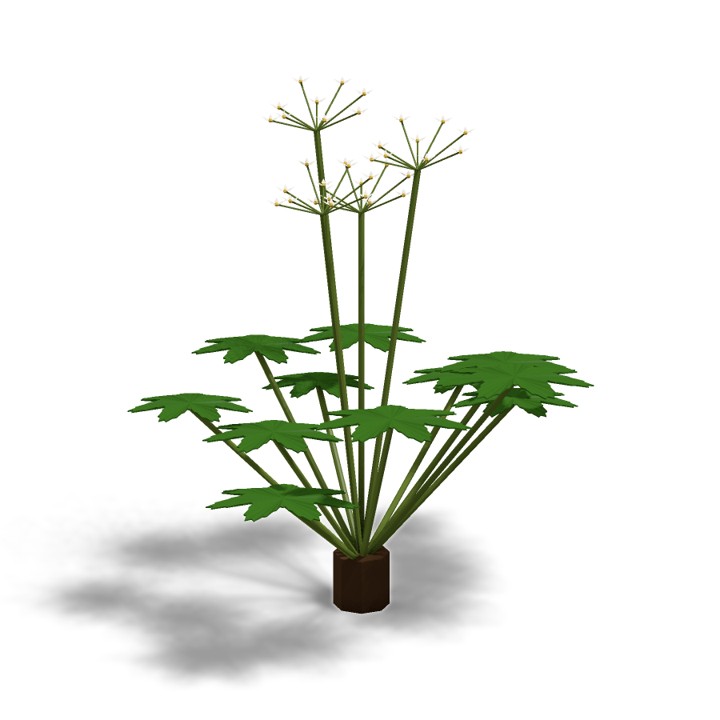} & \PWComparisonImage{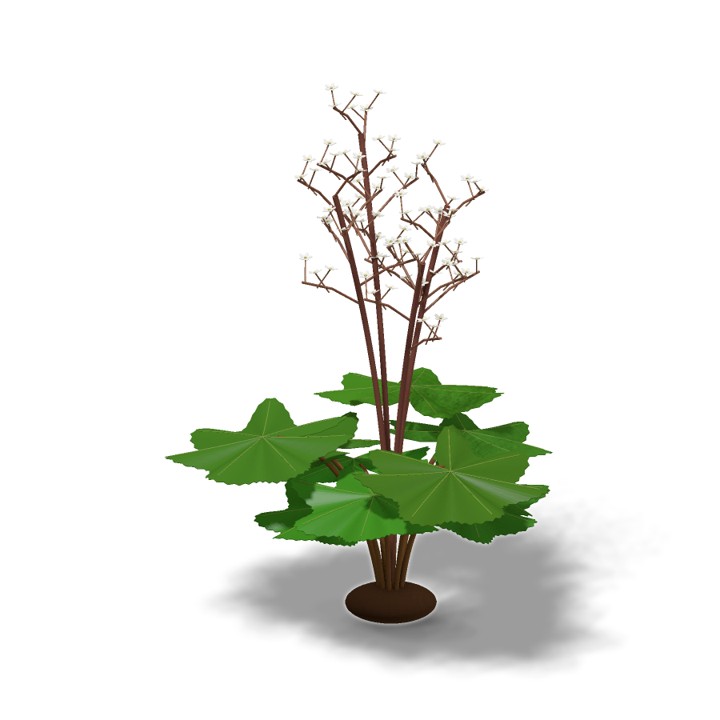} & \PWComparisonImage[Generation failed]{gpt6_astra/mukdenia_floraforge} & \PWComparisonImage{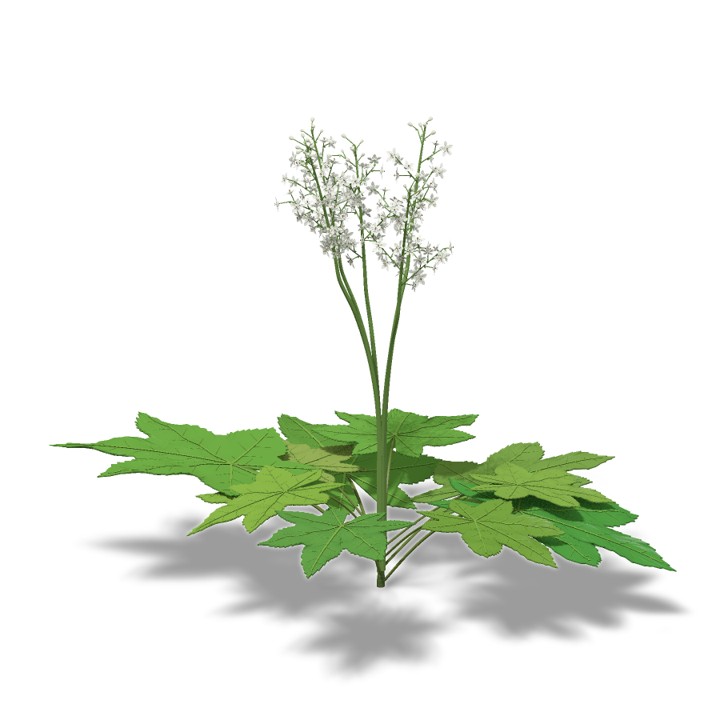} \\[2pt]
    Qwen3.8-27B & \PWComparisonImage{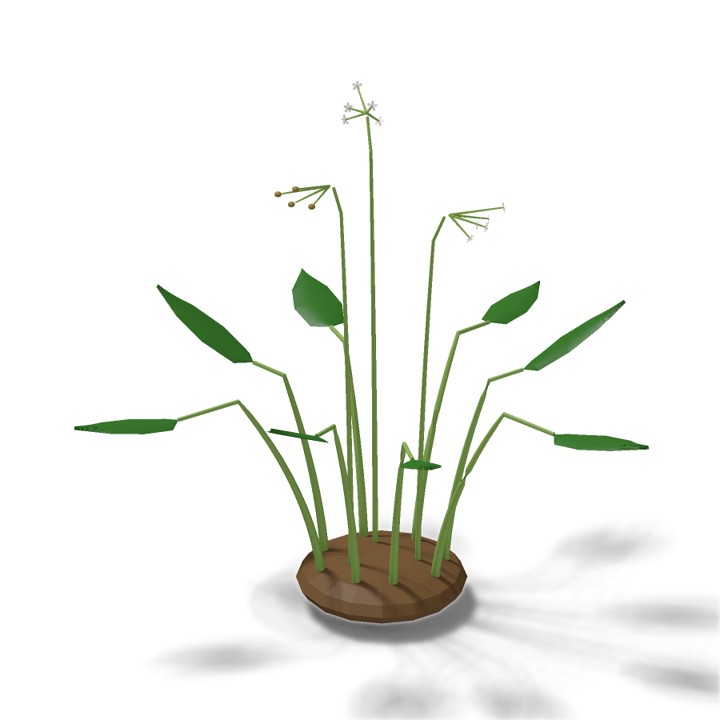} & \PWComparisonImage[Generation failed]{qwen38_27b/mukdenia_deepseek} & \PWComparisonImage{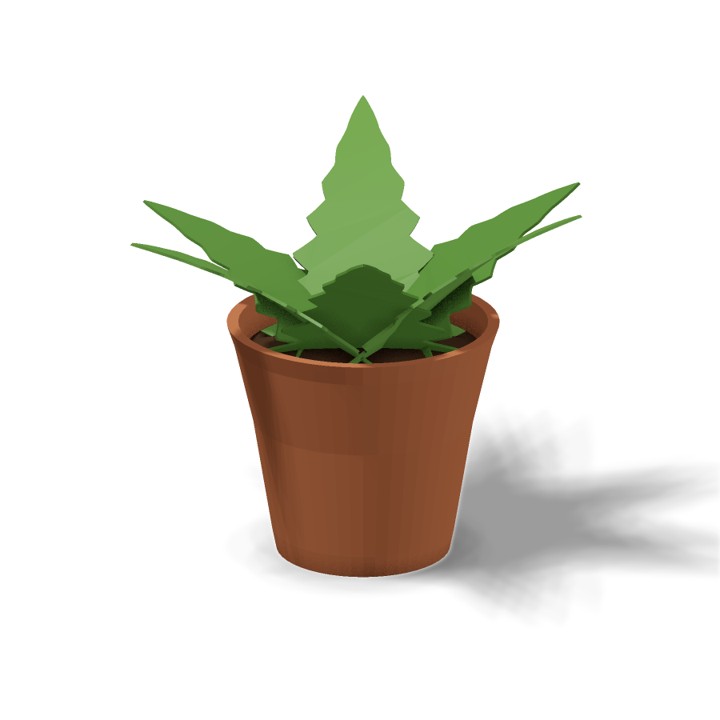} & \PWComparisonImage{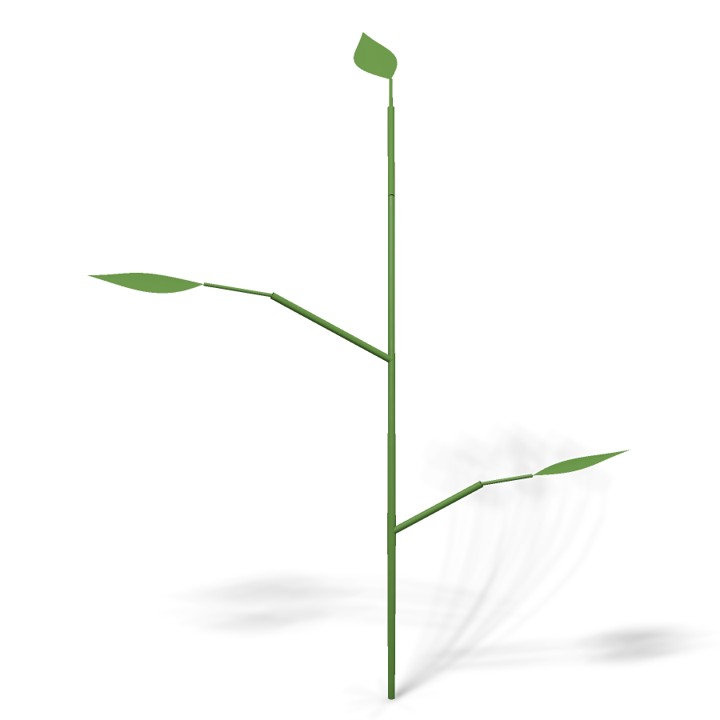} & \PWComparisonImage{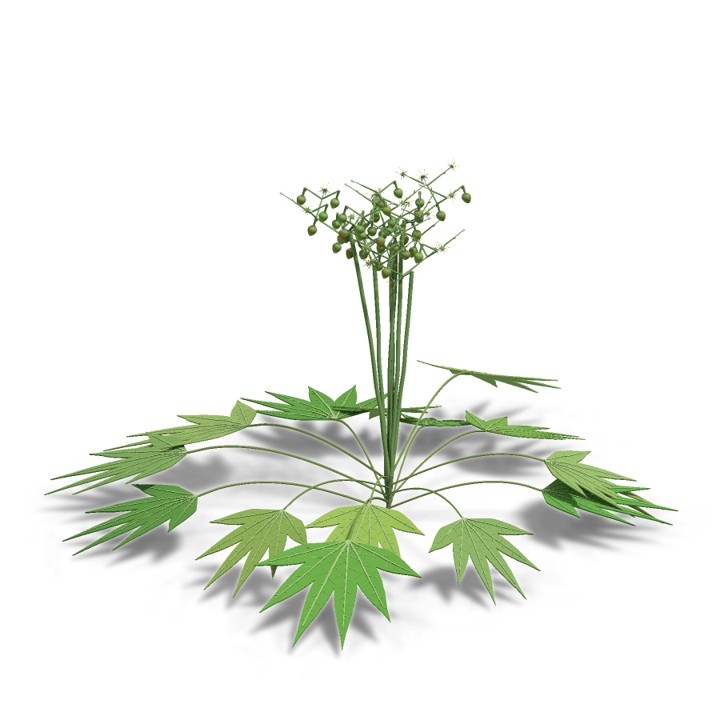} \\
    \bottomrule
  \end{NiceTabular}
  \caption{\textbf{Visual comparison with GPT-6-astra and Qwen3.8-27B.} Each prompt heads two rows of results, one per base LLM; columns compare tailor-made and mass-grown methods, with \ours on the right; insets enlarge a 3\,cm square around the highest open flower, the same for every method. The prompts cover species-level input, numerical constraints, and a less-documented species. \emph{Generation failed}: no asset, or one with no surface that can be drawn.}
  \vspace{5mm}
  \label{fig:comparison}
\end{figure*}

The visual comparison in \cref{fig:comparison} supports these gains across species-level prompts, detailed numerical instructions, and less-documented species. \ours keeps each species' growth form and organ arrangement with both base LLMs, whereas with Qwen3.8-27B several baselines fail to generate an asset or produce simplified plants. Reusing one description for a population thus costs no individual asset quality.

\begin{figure*}[t]
  \centering
  \footnotesize
  \setlength{\tabcolsep}{1.5pt}
  \newlength{\PWPopMainWidth}
  \setlength{\PWPopMainWidth}{\dimexpr(\linewidth-12\tabcolsep)/6\relax}
  \newlength{\PWPopMainHeight}
  \setlength{\PWPopMainHeight}{28mm}
  \newcommand{\PWPopMainBlank}{%
    \fcolorbox{black!18}{black!3}{%
      \parbox[c][\PWPopMainHeight][c]{\dimexpr\PWPopMainWidth-2\fboxsep-2\fboxrule\relax}{%
        \centering\textcolor{red}{\textit{TBD}}%
      }%
    }%
  }
  \newcommand{\PWPopMainRaster}[1]{%
    \includegraphics[width=\PWPopMainWidth,height=\PWPopMainHeight,keepaspectratio]{#1}%
  }
  \newcommand{\PWPopMainCell}[1]{%
    \parbox[c][\dimexpr\PWPopMainHeight+2\fboxsep+2\fboxrule\relax][c]{\PWPopMainWidth}{%
      \centering
      \IfFileExists{figs/individuals/#1.pdf}{\PWPopMainRaster{figs/individuals/#1.pdf}}{%
        \IfFileExists{figs/individuals/#1.png}{\PWPopMainRaster{figs/individuals/#1.png}}{%
          \IfFileExists{figs/individuals/#1.jpg}{\PWPopMainRaster{figs/individuals/#1.jpg}}{%
            \IfFileExists{figs/individuals/#1.jpeg}{\PWPopMainRaster{figs/individuals/#1.jpeg}}{\PWPopMainBlank}%
          }%
        }%
      }%
    }%
  }
  \newcommand{\PWPopMainPrompt}[1]{\multicolumn{6}{l}{\strut\textbf{Prompt:} #1}\\[1pt]}
  \begin{NiceTabular}{*{6}{c}}[cell-space-limits=0pt]
    \toprule
    \PWPopMainPrompt{A garden rose (\emph{Rosa}) bush in flower.}
    {\scriptsize Reviewed during authoring} & \Block{1-5}{\scriptsize Additional individuals from the same description} & & & & \\
    \cmidrule(lr){2-6}
    \PWPopMainCell{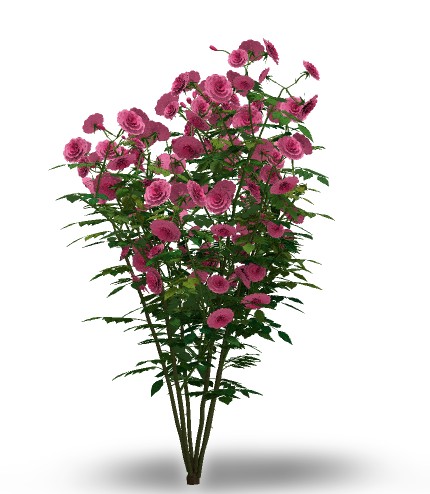} & \PWPopMainCell{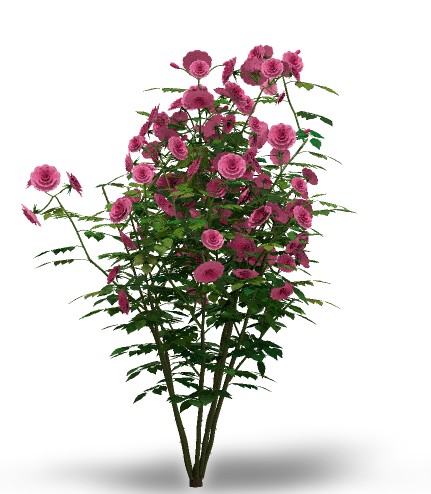} & \PWPopMainCell{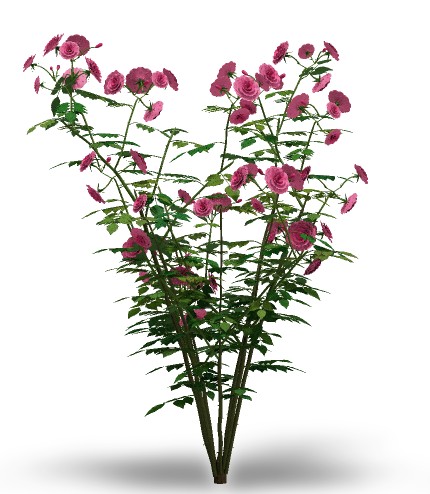} & \PWPopMainCell{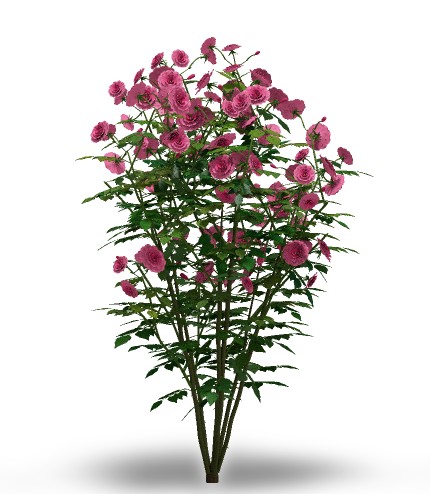} & \PWPopMainCell{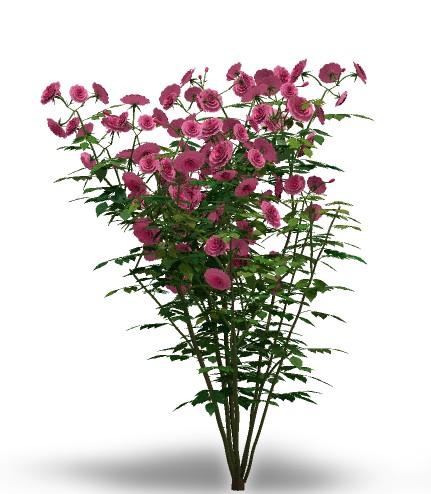} & \PWPopMainCell{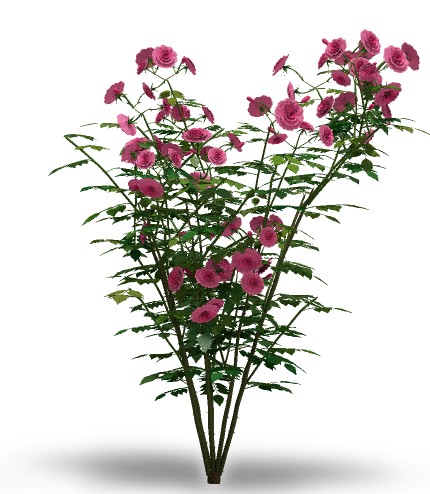} \\
    130\,cm & 125\,cm & 129\,cm & 127\,cm & 122\,cm & 132\,cm \\[2pt]
    \bottomrule
  \end{NiceTabular}\vspace{-2.3mm}%
  \caption{\textbf{Population variation from one description.} Six garden rose individuals at a common scale, with their heights. The leftmost was reviewed during our method; the others required no further LLM inference. Additional species are shown in the supplementary material.}
  \label{fig:individuals}
\end{figure*}

\para{Human preference.}
Raters prefer \ours over Codex in 60.9\% and over DeepSeek Harness in 59.1\% of decisive judgments, and are close to balanced against Articraft (52.8\%), as shown in \cref{tab:human}. A Bradley--Terry fit over all method pairs also ranks \ours first, closely followed by Articraft. \ours is preferred over FloraForge in 63.2\% of decisive judgments; FloraForge builds assets for only two of the 24 species.

\subsection{Ablation study}
\label{sec:exp:lowknow}

\begin{table}[t]
  \centering
  \caption{\textbf{Botanical grounding ablations on GPT-6-astra.} We ablate web search or profiling from the full framework. %
  Bold marks the best scores, including ties.}\vspace{-3mm}
  \label{tab:lowknow}
  \footnotesize
  \setlength{\tabcolsep}{2pt}
  \begin{NiceTabular*}{\linewidth}{@{\hspace{4pt}\extracolsep{\fill}}lccc@{\hspace{4pt}}}[cell-space-limits=1pt]
    \CodeBefore
      \rowcolor{black!4}{1,2}
      \tikz \fill [cyan!6] ([yshift=0.5pt]5-|1) rectangle ([yshift=3.5pt]6-|5) ;
    \Body
    \toprule
     & \Block{1-3}{Botanical correctness} & & \\
    \cmidrule(lr){2-4}
    Setting
      & \shortstack{Well-documented\\(132)}
      & \shortstack{Less-documented\\(50)}
      & \shortstack{All\\(182)} \\
    \midrule
    w/o search    & 113 & 36 & 149 \\
    w/o profiling & 112 & \textbf{41} & 153 \\
    \ours         & \textbf{116} & \textbf{41} & \textbf{157} \\
    \bottomrule
  \end{NiceTabular*}
\end{table}

\para{Settings.}
We assess botanical grounding by removing search or profiling from the full framework with GPT-6-astra (\cref{tab:lowknow}). ``W/o search'' variant constructs the profile from the LLM's knowledge alone. ``W/o profiling'' variant retains search but directly authors the developmental description without a structured profile or its associated checks, including population consistency checks. The developmental vocabulary, growth engine, compiler, remaining checks, and revision budget are unchanged.

\para{Results.}
Removing search reduces botanical correctness for both well-documented species (116 to 113) and less-documented species (41 to 36), showing that its benefit extends beyond species with limited online information. Removing profiling lowers agreement on well-documented species from 116 to 112, while agreement on less-documented species remains at 41. Profiling thus adds agreement only on the well-documented species, where search returns more information to organize.

\subsection{Population generation}
\label{sec:exp:population}

\begin{table}[t]
  \centering
  \caption{\textbf{Population consistency and generation cost.} Profile consistency is the percentage of evaluated individuals whose checked traits all agree with their botanical profile. For each finalized description, we compare 100 seeds used for validation during authoring with 100 unseen seeds. Initial-asset times include research, authoring, validation, and compilation; times are medians.}
  \label{tab:models}
  \footnotesize
  \setlength{\tabcolsep}{2pt}
  \begin{NiceTabular*}{\linewidth}{@{\hspace{4pt}\extracolsep{\fill}}lcc@{\hspace{4pt}}}[cell-space-limits=1pt]
    \CodeBefore
      \rowcolor{black!4}{1,2}
    \Body
    \toprule
     & \Block{1-2}{Profile consistency (\%)} & \\
    \cmidrule(lr){2-3}
    Base LLM & During authoring & Additional individuals \\
    \midrule
    GPT-6-astra & 95.1 & 94.2 \\
    GPT-5.5 & 99.8 & 99.4 \\
    Qwen3.8-27B & 53.2 & 52.6 \\
    \bottomrule
  \end{NiceTabular*}
  \par\medskip
  \begin{NiceTabular*}{\linewidth}{@{\hspace{4pt}\extracolsep{\fill}}lcc@{\hspace{4pt}}}[cell-space-limits=1pt]
    \CodeBefore
      \rowcolor{black!4}{1,2}
    \Body
    \toprule
     & \Block{1-2}{Generation time} & \\
    \cmidrule(lr){2-3}
    Base LLM & \shortstack{Initial asset\\(minutes)}
      & \shortstack{Additional individual\\(seconds)} \\
    \midrule
    GPT-6-astra & 12 & 47 \\
    GPT-5.5 & 15 & 65 \\
    Qwen3.8-27B & 65 & 30 \\
    \bottomrule
  \end{NiceTabular*}
  \vspace{-3mm}
\end{table}

\para{Settings.}
We evaluate each finalized developmental description on two sets of 100 seeds: those used for population-level validation during authoring and previously unseen seeds. We apply the same programmatic trait checks to both sets of growth traces against the finalized botanical profile, reporting the percentage of individuals for which all checked traits agree as the profile consistency. \Cref{fig:individuals} illustrates individual variation from a shared description; additional species are shown in the supplementary material.

\para{Consistency across individuals.}
Profile consistency changes from 95.1\% during authoring to 94.2\% on unseen seeds with GPT-6-astra, from 99.8\% to 99.4\% with GPT-5.5, and from 53.2\% to 52.6\% with Qwen3.8-27B (\cref{tab:models}). The decrease stays below one percentage point for every base LLM, indicating little loss of consistency when sampling individuals. For GPT-6-astra, 13 of 22 descriptions satisfy all profile checks for every additional individual.

\para{Generation cost.}
Generating the first asset, including research, authoring, validation, and compilation, takes a median of 12 minutes with GPT-6-astra, 15 minutes with GPT-5.5, and 65 minutes with Qwen3.8-27B. Each additional individual reuses the finalized description and requires only growth and compilation, without further LLM inference: a median of 47, 65, and 30 seconds, respectively (\cref{tab:models}). Tailor-made methods instead repeat LLM authoring for each asset. With GPT-6-astra, Codex takes a median of 3.3 minutes and DeepSeek Harness 1.4 minutes per asset, so a population of 100 individuals takes \ours about 1.5 hours, against 5.6 hours for Codex and 2.3 hours for DeepSeek Harness. With Qwen3.8-27B, a single Codex asset takes a median of 4.5 hours, including runs stopped at the time limit.

\section{Conclusions}
\label{sec:conclusion}

We presented \ours, an agentic framework that translates text prompts into reusable developmental descriptions and populations of simulation-ready plants. A botanically grounded LLM agent researches, authors, validates, and revises these descriptions, while a shared growth engine and compiler preserve organ identities and connectivity through generation and physical instantiation. Experiments show improved botanical correctness across base LLMs and comparable or better human preference than tailor-made baselines. Once a description is authored, additional individuals require no further LLM inference, so populations for agricultural perception and robot simulation can be generated at scale.

\para{Limitations.}
Botanical fidelity depends on source quality, LLM interpretation, and the developmental vocabulary; profile-based checks may retain incorrect traits or miss inconsistencies across individuals. Our mechanical models and physical traits are approximations, and agreement with measured plant mechanics and population variation remains unvalidated.

\section*{Acknowledgments}
This work was partly supported by JSPS KAKENHI JP23H05491, JST CREST JPMJCR26E1, and JST ASPIRE JPMJAP2502.

{
    \small
    \bibliographystyle{ieeenat_fullname}
    \bibliography{main}
}

\clearpage
\setcounter{page}{1}
\maketitlesupplementary
\appendix

\section{Prompts}
\label{sec:supp:prompts}

\paragraph{Development.}
We developed the method on several sets of prompts, among them the ten crop and garden plants of
\cref{tab:supp:ten}, the twenty-nine plants of nine growth forms of \cref{tab:supp:thirty}, twelve
complex prompts that each state 5 to 12 checkable constraints (ripeness, open flowers, training, organ forms), six species little documented on the web, twenty species whose floras use many of the organ terms of the vocabulary, and the twenty-four crop and garden plants of the human study (\cref{sec:exp:survey}). Four of the twenty-nine are stress tests that ask for more weight than a stem of the usual stiffness can carry.

\paragraph{Test.}
\Cref{tab:supp:test} lists the test set (\cref{sec:exp:setup}). No test prompt is among the
prompts run during development. Each names the plant and its developmental stage, including the six less-documented species, which have a treatment in the Flora of China but no English Wikipedia article or only a stub. \Cref{sec:supp:reference} describes the reference traits and how they are scored.

\begin{table*}[t]
  \centering
  \caption{\textbf{The ten crop and garden plants.}}
  \label{tab:supp:ten}
  \footnotesize
  \begin{tabular}{@{}lp{0.78\linewidth}@{}}
  \toprule
  Plant & Prompt \\
  \midrule
    Greenhouse tomato & A greenhouse tomato plant about 80 cm tall on a single upright stem without side branches, with compound leaves and 3 fruit trusses of 4 to 6 round red tomatoes each. \\
    Bell pepper & A bell pepper plant about 50 cm tall whose main stem forks into two branches, with simple oval leaves and 3 large green bell peppers. \\
    Eggplant & An eggplant plant about 70 cm tall with large simple leaves and 4 long, dark purple eggplants hanging from it. \\
    Strawberry & A strawberry plant: a low rosette about 20 cm tall with about 10 trifoliate leaves on long stalks and 5 red strawberries hanging on thin stems. \\
    Soybean seedling & A young soybean seedling about 25 cm tall on one upright stem with 4 trifoliate leaves arranged alternately, with no flowers or pods yet. \\
    Basil & A basil plant about 30 cm tall with bright green, simple oval leaves in opposite pairs, 6 pairs along the main stem, and no flowers. \\
    Sunflower & A single sunflower 1.5 m tall with one large yellow flower head at the top and about 14 large simple leaves alternating up the stem. \\
    Cherry tomato & A cherry tomato plant with 2 long trusses, each carrying about 12 small round red cherry tomatoes. \\
    Bush bean & A bush bean plant about 40 cm tall with trifoliate leaves, a few small white flowers, and 8 long, slender green pods. \\
    Chili pepper & A chili pepper plant about 40 cm tall, bushy with several side branches, narrow pointed simple leaves, and 10 small red chili peppers hanging downward. \\
  \bottomrule
\end{tabular}

\end{table*}

\begin{table*}[t]
  \centering
  \caption{\textbf{The twenty-nine plants of nine growth forms.}}
  \label{tab:supp:thirty}
  \scriptsize
  \begin{tabular}{@{}llp{0.66\linewidth}@{}}
  \toprule
  Plant & Form & Prompt \\
  \midrule
    wheat & cereal & A wheat plant about 80 cm tall with 4 stems from its base (the main stem and 3 tillers), each ending in one upright ear, and long narrow leaves. \\
    rice & cereal & A rice plant about 90 cm tall with 5 stems from its base, each ending in one drooping panicle of grains, and long narrow leaves. \\
    cucumber & vine & A cucumber vine about 1.5 m tall climbing upward, with large lobed simple leaves, curling tendrils and 4 long green cucumbers. \\
    pea & vine & A pea plant about 60 cm tall that climbs with tendrils, with compound leaves, a few white flowers and 6 green pods. \\
    apple & tree & A young apple tree about 1.5 m tall with one trunk and 4 main side branches, simple oval leaves and 6 red apples. \\
    lemon & tree & A potted lemon tree about 1 m tall whose short trunk branches into a bushy crown, with glossy simple leaves and 5 yellow lemons. \\
    lettuce & rosette & A lettuce plant: a loose rosette about 20 cm tall and 30 cm across with about 16 large frilly leaves and no flower stalk. \\
    spinach & rosette & A spinach plant about 25 cm tall: a rosette of about 12 dark green oval leaves on long stalks. \\
    zucchini & cucurbit & A zucchini plant about 60 cm tall, a compact bush with very large lobed leaves, 3 yellow flowers and 2 long green zucchinis. \\
    cotton & field crop & A cotton plant about 1 m tall with 3 side branches, lobed leaves and 8 open white cotton bolls. \\
    mint & herb & A mint plant about 30 cm tall with 3 upright stems, opposite pairs of small toothed leaves, and no flowers. \\
    rose & ornamental & A rose bush about 80 cm tall with 4 canes, compound leaves of 5 leaflets, and 3 red flowers. \\
    sunflower heavy & stress & A sunflower 2 m tall on a slender stem, with one very heavy seed head 35 cm across that makes the top nod, and about 16 large simple leaves. \\
    tomato overloaded & stress & A young tomato plant only 40 cm tall on one slender stem, carrying a single truss of 5 large beefsteak tomatoes. \\
    pepper overloaded & stress & A small pepper seedling about 30 cm tall on a thin single stem, weighed down by 4 large bell peppers. \\
    hollyhock & stress & A hollyhock 2 m tall: a single thin flowering stalk with about 20 large flowers along its upper half and large rounded leaves below. \\
    maize & cereal & A maize plant about 2 m tall on a single stalk with about 14 long strap-shaped leaves alternating up it, one tassel at the top and 2 ears lower down. \\
    barley & cereal & A barley plant about 70 cm tall with 6 stems from its base, each ending in one nodding bearded ear. \\
    grape & vine & A grapevine shoot about 1.2 m long with large lobed leaves, tendrils, and 2 hanging bunches of about 40 small purple grapes each. \\
    runner bean & vine & A runner bean vine about 2 m tall twining upward, with trifoliate leaves, red flowers and 10 long green pods. \\
    peach & tree & A peach tree about 2 m tall with 3 main branches spreading from the trunk, narrow pointed leaves and 8 round peaches. \\
    blueberry & tree & A blueberry bush about 1 m tall with 6 upright woody canes from its base, small oval leaves and about 30 blue berries in small clusters. \\
    celery & rosette & A celery plant about 45 cm tall: a clump of about 10 long upright ribbed stalks, each ending in a compound leaf. \\
    bok choy & rosette & A bok choy plant about 30 cm tall with about 10 thick white leaf stalks and broad dark green leaf blades. \\
    melon & cucurbit & A melon vine trailing about 2 m along the ground, with lobed leaves, tendrils and 2 round netted melons. \\
    chickpea & field crop & A chickpea plant about 50 cm tall, bushy, with small compound leaves and 12 small inflated pods. \\
    okra & field crop & An okra plant about 1.2 m tall on one upright stem with large lobed leaves and 5 long green pods pointing upward. \\
    tulip & ornamental & A tulip about 40 cm tall with 3 broad leaves at its base and one red cup-shaped flower on a single stem. \\
    potato & field crop & A potato plant about 60 cm tall with 3 stems, compound leaves of 7 leaflets, and 5 white flowers. \\
  \bottomrule
\end{tabular}

\end{table*}

\begin{table}[t]
  \centering
  \caption{\textbf{The test set}: 22 species prompts in three groups (A: species of growth forms the
vocabulary covers; B: growth forms it was not built for; C: less-documented species), with the number of reference traits scored for each.}
  \label{tab:supp:test}
  \scriptsize
  \begin{tabular}{@{}lp{0.74\linewidth}c@{}}
  \toprule
  Group & Prompt & Traits \\
  \midrule
  A & A young avocado (\emph{Persea americana}) tree with fruit & 13 \\
  A & A blackcurrant (\emph{Ribes nigrum}) bush with ripe berries & 14 \\
  A & A young fig (\emph{Ficus carica}) tree with figs & 13 \\
  A & A pomegranate (\emph{Punica granatum}) shrub with flowers and fruit & 14 \\
  A & A flowering garden peony (\emph{Paeonia lactiflora}) plant & 12 \\
  A & A flowering peanut (\emph{Arachis hypogaea}) plant & 12 \\
  A & A flowering gardenia (\emph{Gardenia jasminoides}) shrub & 15 \\
  A & A sorghum (\emph{Sorghum bicolor}) plant with a ripe panicle & 11 \\
  A & A young olive (\emph{Olea europaea}) tree with fruit & 13 \\
  A & A flowering rosemary (\emph{Salvia rosmarinus}) shrub & 13 \\
  \midrule
  B & A cabbage (\emph{Brassica oleracea} var.~\emph{capitata}) plant with a firm head & 11 \\
  B & A pineapple (\emph{Ananas comosus}) plant with a ripening fruit & 14 \\
  B & A young Scots pine (\emph{Pinus sylvestris}) tree with cones & 10 \\
  B & A young date palm (\emph{Phoenix dactylifera}) with a bunch of dates & 10 \\
  B & A papaya (\emph{Carica papaya}) tree with ripening fruit & 15 \\
  B & An asparagus (\emph{Asparagus officinalis}) plant with red berries & 10 \\
  \midrule
  C & A flowering \emph{Asteropyrum peltatum} plant & 13 \\
  C & A flowering \emph{Beesia calthifolia} plant & 12 \\
  C & A flowering \emph{Dysosma versipellis} plant & 15 \\
  C & A flowering \emph{Mukdenia rossii} plant & 12 \\
  C & A flowering \emph{Peltoboykinia tellimoides} plant & 12 \\
  C & A flowering \emph{Ypsilandra thibetica} plant & 12 \\
  \bottomrule
\end{tabular}

\end{table}

\paragraph{Detailed prompts.}
\Cref{tab:supp:specified} lists the detailed prompt for each test species (\cref{tab:astra}),
written and registered before any run. A number written exactly is an exact target and ``about $N$'' allows 20\% either way; every number a prompt states is measured, words such as ``ripe'' are not. The stated attributes are the plant's height, its stems from the base, its main branches (shoots from the main stem), its side shoots, leaves, flowers, fruit (a pine's cones) and inflorescences (a bunch or a cluster).

\begin{table}[t]
  \centering
  \caption{\textbf{The detailed prompts}, one per species of the test set, with the
number of attributes each states (66 in all).}
  \label{tab:supp:specified}
  \scriptsize
  \begin{tabular}{@{}lp{0.74\linewidth}c@{}}
  \toprule
  Group & Prompt & Stated \\
  \midrule
  A & An avocado (\emph{Persea americana}) tree about 2.5 m tall with 3 main branches and 6 fruit & 3 \\
  A & A blackcurrant (\emph{Ribes nigrum}) bush about 1.2 m tall with 6 stems from its base and 5 clusters of ripe berries & 3 \\
  A & A fig (\emph{Ficus carica}) tree about 2 m tall with 4 main branches and 10 figs & 3 \\
  A & A pomegranate (\emph{Punica granatum}) shrub about 1.5 m tall with 4 flowers and 3 fruit & 3 \\
  A & A flowering garden peony (\emph{Paeonia lactiflora}) plant about 80 cm tall with 5 stems and 5 flowers & 3 \\
  A & A flowering peanut (\emph{Arachis hypogaea}) plant about 40 cm tall with 4 main branches and about 12 flowers & 3 \\
  A & A flowering gardenia (\emph{Gardenia jasminoides}) shrub about 1 m tall with 4 main branches and 8 flowers & 3 \\
  A & A sorghum (\emph{Sorghum bicolor}) plant about 2 m tall on a single stem with about 12 leaves and one ripe panicle & 4 \\
  A & An olive (\emph{Olea europaea}) tree about 1.8 m tall with 3 main branches and about 20 fruit & 3 \\
  A & A flowering rosemary (\emph{Salvia rosmarinus}) shrub about 60 cm tall with 6 main branches and about 40 flowers & 3 \\
  \midrule
  B & A cabbage (\emph{Brassica oleracea} var.~\emph{capitata}) plant about 35 cm tall with a firm head and no flowers & 2 \\
  B & A pineapple (\emph{Ananas comosus}) plant about 1 m tall with about 30 leaves and one ripening fruit & 3 \\
  B & A Scots pine (\emph{Pinus sylvestris}) tree about 3 m tall with 12 side branches and 6 cones & 3 \\
  B & A date palm (\emph{Phoenix dactylifera}) about 4 m tall with 12 leaves and one bunch of dates & 3 \\
  B & A papaya (\emph{Carica papaya}) tree about 2.5 m tall with about 15 leaves and 8 ripening fruit & 3 \\
  B & An asparagus (\emph{Asparagus officinalis}) plant about 1.2 m tall with 4 stems and about 20 red berries & 3 \\
  \midrule
  C & A flowering \emph{Asteropyrum peltatum} plant about 12 cm tall with 4 leaves and 2 flowers & 3 \\
  C & A flowering \emph{Beesia calthifolia} plant about 40 cm tall with 3 basal leaves and one flowering stem & 3 \\
  C & A flowering \emph{Dysosma versipellis} plant about 30 cm tall with 2 leaves and 5 flowers & 3 \\
  C & A flowering \emph{Mukdenia rossii} plant about 30 cm tall with 5 basal leaves and 2 flowering stems & 3 \\
  C & A flowering \emph{Peltoboykinia tellimoides} plant about 50 cm tall with 3 leaves and one flowering stem & 3 \\
  C & A flowering \emph{Ypsilandra thibetica} plant about 25 cm tall with 8 basal leaves and 2 flowering stems & 3 \\
  \bottomrule
\end{tabular}

\end{table}

\section{Additional population examples}
\label{sec:supp:population}

\begin{figure*}[t]
  \centering
  \footnotesize
  \setlength{\tabcolsep}{1.5pt}
  \newlength{\PWPopSuppWidth}
  \setlength{\PWPopSuppWidth}{\dimexpr(\linewidth-12\tabcolsep)/6\relax}
  \newlength{\PWPopSuppHeight}
  \setlength{\PWPopSuppHeight}{28mm}
  \newcommand{\PWPopSuppBlank}{%
    \fcolorbox{black!18}{black!3}{%
      \parbox[c][\PWPopSuppHeight][c]{\dimexpr\PWPopSuppWidth-2\fboxsep-2\fboxrule\relax}{%
        \centering\textcolor{red}{\textit{TBD}}%
      }%
    }%
  }
  \newcommand{\PWPopSuppRaster}[1]{%
    \includegraphics[width=\PWPopSuppWidth,height=\PWPopSuppHeight,keepaspectratio]{#1}%
  }
  \newcommand{\PWPopSuppCell}[1]{%
    \parbox[c][\dimexpr\PWPopSuppHeight+2\fboxsep+2\fboxrule\relax][c]{\PWPopSuppWidth}{%
      \centering
      \IfFileExists{figs/individuals/#1.pdf}{\PWPopSuppRaster{figs/individuals/#1.pdf}}{%
        \IfFileExists{figs/individuals/#1.png}{\PWPopSuppRaster{figs/individuals/#1.png}}{%
          \IfFileExists{figs/individuals/#1.jpg}{\PWPopSuppRaster{figs/individuals/#1.jpg}}{%
            \IfFileExists{figs/individuals/#1.jpeg}{\PWPopSuppRaster{figs/individuals/#1.jpeg}}{\PWPopSuppBlank}%
          }%
        }%
      }%
    }%
  }
  \newcommand{\PWPopSuppPrompt}[1]{\multicolumn{6}{l}{\strut\textbf{Prompt:} #1}\\[1pt]}
  \begin{NiceTabular}{*{6}{c}}[cell-space-limits=0pt]
    \toprule
    {\scriptsize Reviewed during authoring} & \Block{1-5}{\scriptsize Additional individuals from the same description} & & & & \\
    \cmidrule(lr){2-6}
    \PWPopSuppPrompt{A maize (Zea mays) plant with a tassel and ears.}
    \PWPopSuppCell{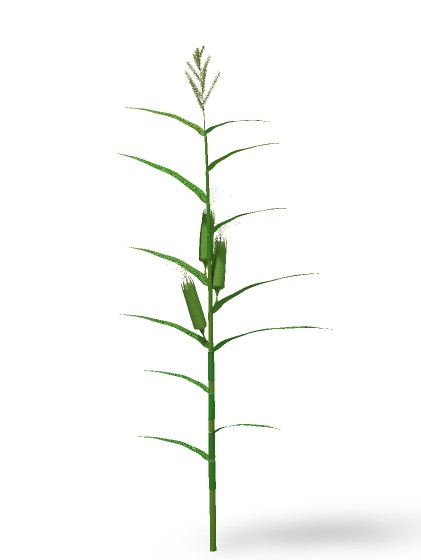} & \PWPopSuppCell{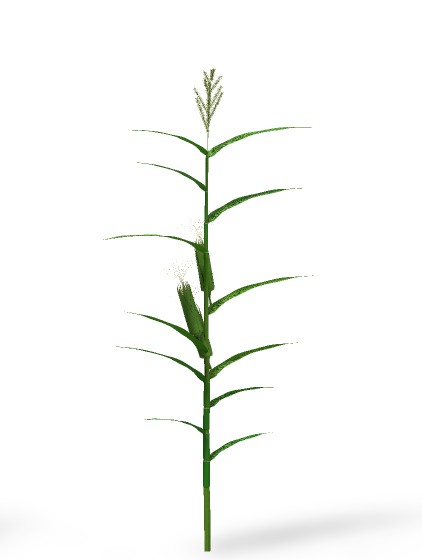} & \PWPopSuppCell{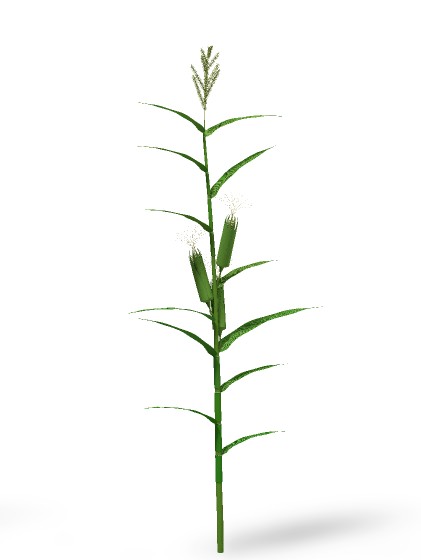} & \PWPopSuppCell{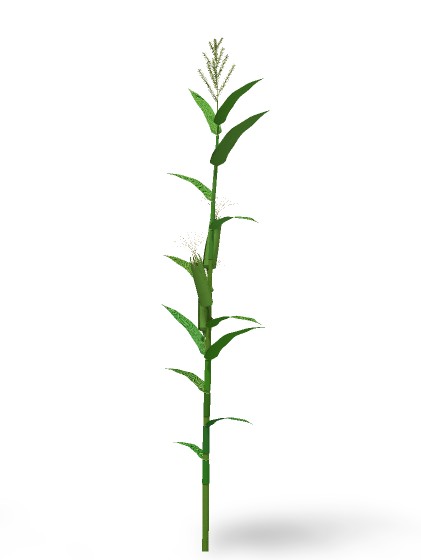} & \PWPopSuppCell{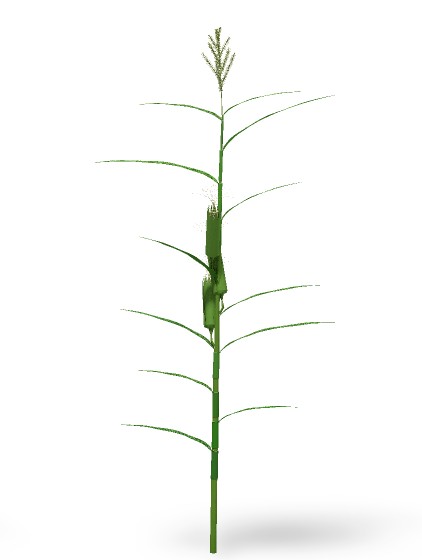} & \PWPopSuppCell{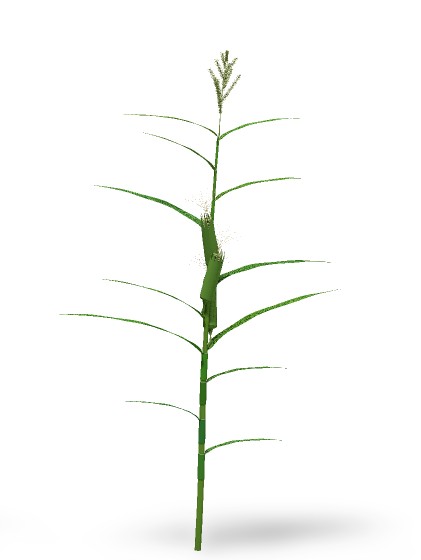} \\
    257\,cm, 2.32\,kg & 246\,cm, 2.28\,kg & 258\,cm, 2.32\,kg & 262\,cm, 2.38\,kg & 267\,cm, 2.33\,kg & 256\,cm, 2.30\,kg \\[2pt]
    \midrule
    \PWPopSuppPrompt{A flowering Chinese hibiscus (Hibiscus rosa-sinensis) shrub.}
    \PWPopSuppCell{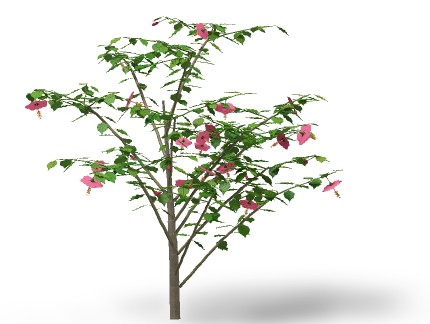} & \PWPopSuppCell{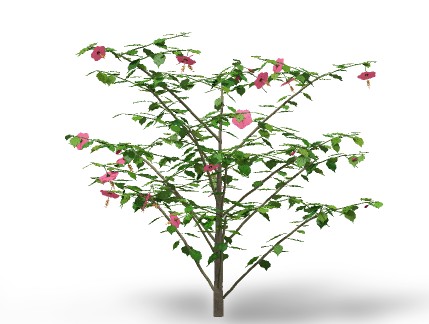} & \PWPopSuppCell{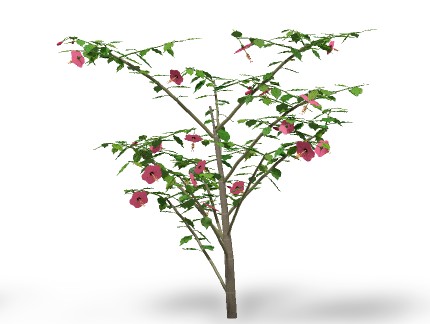} & \PWPopSuppCell{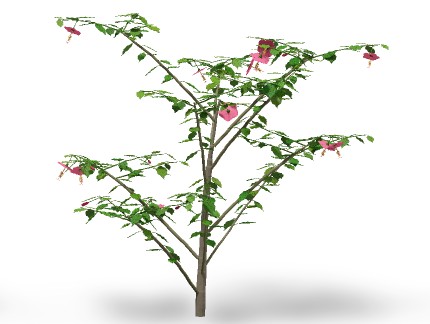} & \PWPopSuppCell{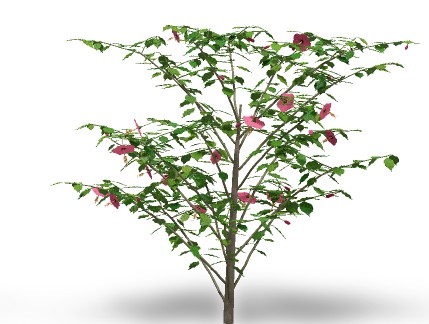} & \PWPopSuppCell{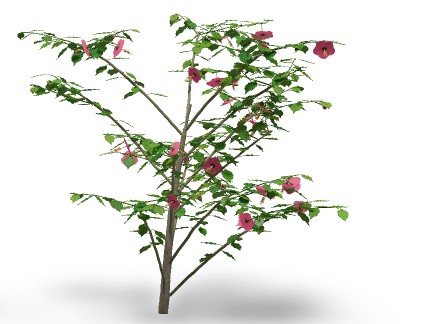} \\
    156\,cm, 3.34\,kg & 152\,cm, 3.34\,kg & 154\,cm, 3.09\,kg & 160\,cm, 3.08\,kg & 156\,cm, 3.86\,kg & 155\,cm, 3.47\,kg \\[2pt]
    \midrule
    \PWPopSuppPrompt{A corn poppy (Papaver rhoeas) plant with red flowers, nodding buds and seed capsules.}
    \PWPopSuppCell{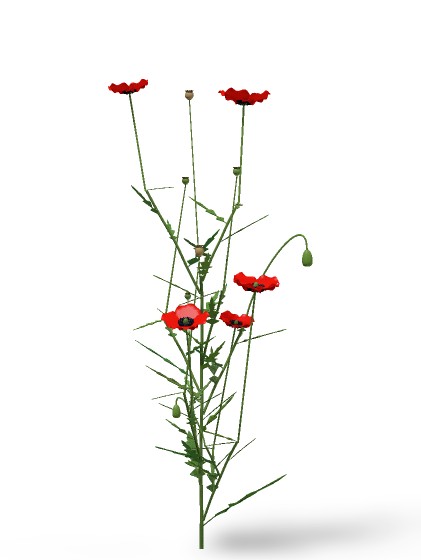} & \PWPopSuppCell{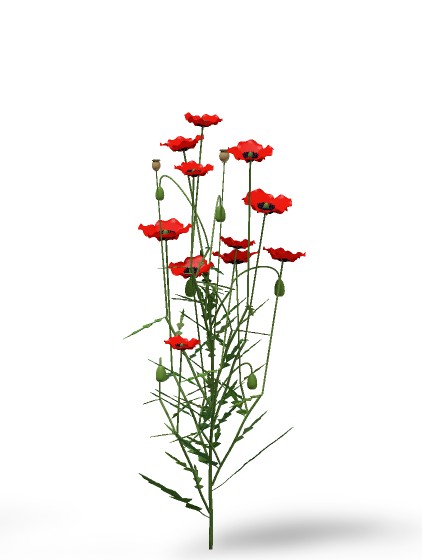} & \PWPopSuppCell{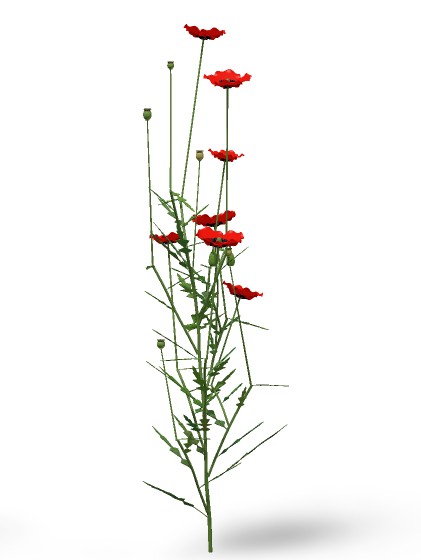} & \PWPopSuppCell{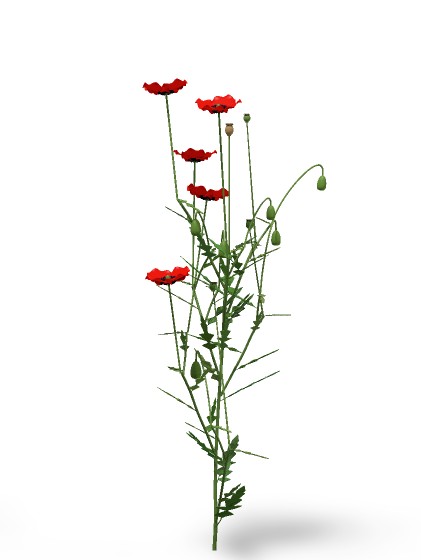} & \PWPopSuppCell{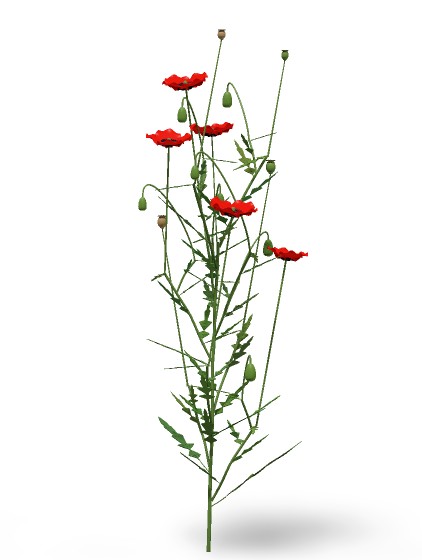} & \PWPopSuppCell{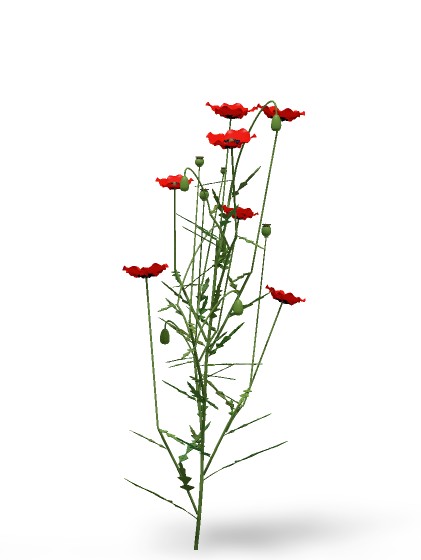} \\
    69\,cm, 0.05\,kg & 64\,cm, 0.09\,kg & 76\,cm, 0.06\,kg & 69\,cm, 0.07\,kg & 76\,cm, 0.07\,kg & 66\,cm, 0.07\,kg \\[2pt]
    \bottomrule
  \end{NiceTabular}
  \caption{\textbf{Additional population examples.} Six individuals from each of three developmental descriptions, shown at a common scale within each row with their heights and masses. Only the leftmost individual of each row was visually reviewed; the others were generated without further LLM inference. These illustrative species are separate from the 22-species quantitative evaluation.}
  \label{fig:supp:individuals}
\end{figure*}

\Cref{fig:supp:individuals} shows population examples of three more species, none of them in the 22-species quantitative test set; the main paper shows the garden rose (\cref{fig:individuals}).

A second set of unseen seeds gives profile consistency within one percentage point of the first (\cref{tab:models}). Over seeds 201 to 300 of every final test-set description, 94.9\% of the individuals agree with their profile in every checked trait with GPT-6-astra, 99.7\% with GPT-5.5, and 53.0\% with Qwen3.8-27B, against 94.2\%, 99.4\%, and 52.6\% over seeds 101 to 200. No description differs between the two sets by more than nine of its hundred individuals.

\section{Human study}
\label{sec:supp:survey}

\paragraph{Rating protocol.}
Raters work through a web page, one pair per page. Each page states the prompt both systems were
given and shows photographs of the real plant beside a video of the two plants turning side by side,
A on the left and B on the right. Four questions follow, on stems and branches, leaves, flowers and
fruit. Each lists what floras and encyclopedias say about that part, every statement with its source,
and shows a photograph of the real part, where one is available, beside close-ups of the part on A
and on B. A fifth question asks about the plant as a whole. The answers are ``A is closer'', ``About
the same'' and ``B is closer'' to the real plant; the part questions also offer ``Not applicable''
when neither plant has the part. Raters are told to ignore size, since each plant is scaled to its
view, and, where a prompt names no flower color, to compare the flowers' form and posture rather
than their color. Across the plants, each system of a pair appears on the left about as often as on
the right. The pages cycle through the plants one pair at a time, and a plant's pairs are ordered so
that consecutive pairs share no system where possible. Each rater judges every pair, at a median of
30\,s a pair; \cref{tab:human} reports the overall question.

\paragraph{Our plants.}

All twenty-four plants our method made for the human-study set (\cref{sec:exp:survey}) passed every check in the final round. Across individuals 101 to 200 of each description (\cref{sec:exp:population}), 2374 of the 2379 that grow agree with its profile in every trait, at least 98\% in every description; 21 of the blueberry's individuals exceed the engine's growth limit. With every hinge nudged in MuJoCo (\cref{sec:supp:impl}), 22 of the plants return and the sunflower is returning at the end of 15\,s; the petunia folds. The coding agents' stages each hold several articulated objects, which the simulator translation we use does not load, so they are not nudged.

\paragraph{Agreement among raters.}
On the overall question, Krippendorff's $\alpha$ over the 152 rated pairs is 0.20 (0.22 when the
answers are treated as ordered), and two raters who judged the same pair gave the same answer in
59\% of cases. Agreement is higher on flowers (0.31) and fruit (0.32) than on stems (0.19) and
leaves (0.21). \Cref{tab:human} pools the judgments of all raters.

\newpage\mbox{}\newpage\mbox{}\newpage  
\section{The tomato under three supports}
\label{sec:supp:support}

\begin{figure}[t]
  \centering
  \footnotesize
  \setlength{\tabcolsep}{2pt}
  \newcommand{\cell}[1]{\includegraphics[height=0.42\linewidth]{figs/support_tomato/#1}}
  \begin{tabular}{@{}ccc@{}}
    \textbf{Self-supporting} & \textbf{Staked} & \textbf{Sprawling} \\[2pt]
    \cell{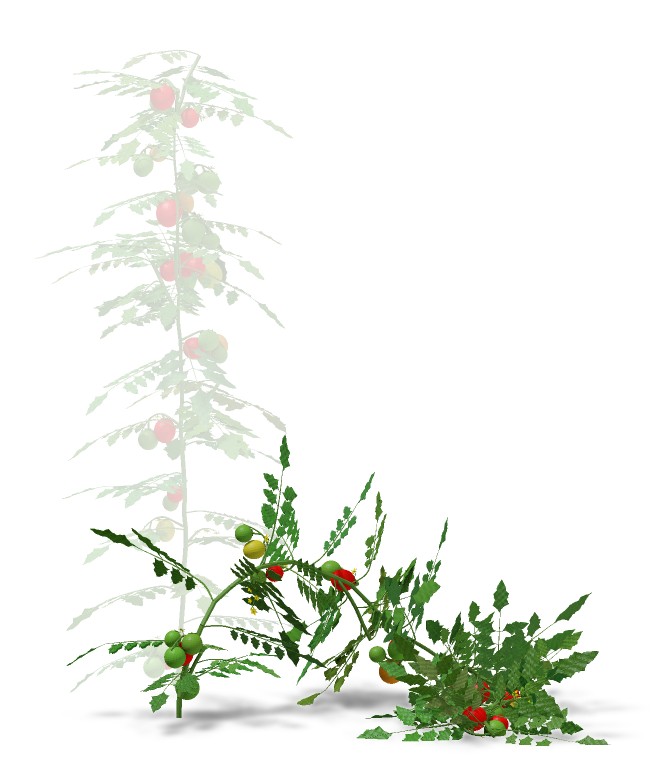} & \cell{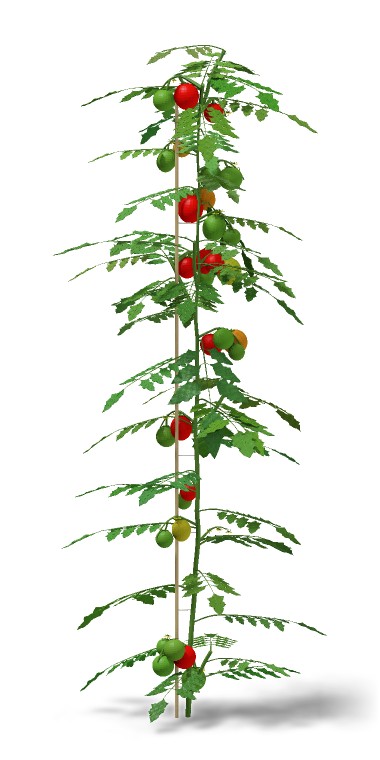} & \cell{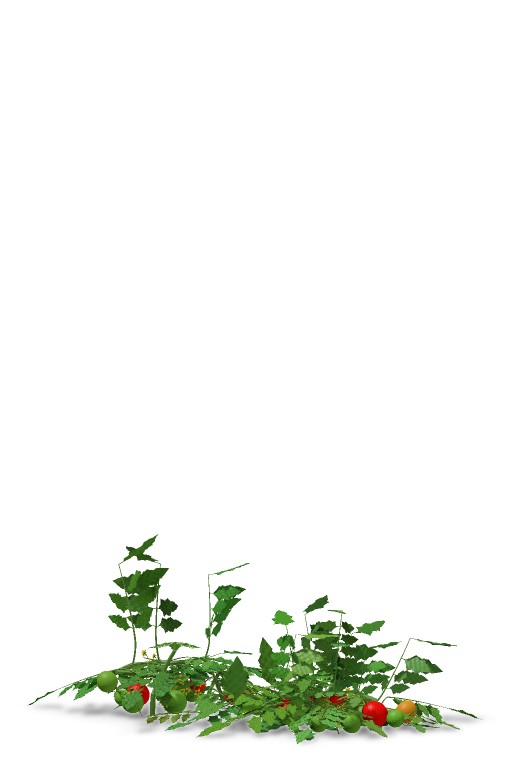} \\
  \end{tabular}
  \caption{\textbf{One tomato under three support habits.} The greenhouse tomato of the
human-study set, which the model staked, seed~0 under each support, drawn to one scale. Self-supporting, its stems cannot carry it even stiffened ten times: drawn where MuJoCo leaves it after 15\,s, its drawn pose faint behind. Staked, its main stem stands rigid beside the stake. Sprawling, it settles under its own weight onto the ground, 370 hinges giving way, its leaves held on their stalks.}
  \label{fig:supp:support}
\end{figure}

On the human-study set, the model staked the greenhouse tomato and the pea that climbs by tendrils, twined the morning glory up a pole, let the strawberry and the petunia sprawl, and let the other nineteen plants stand on their own. \Cref{fig:supp:support} shows the tomato under each of three supports. Self-supporting, it cannot hold itself up: with its weakest hinges stiffened ten times it still folds at five, and the model is told so; in MuJoCo it folds by up to 1.45\,rad within 2\,s. Staked, it stands with two hinges stiffened by 1.5\%, and in MuJoCo it holds its pose for 15\,s to within 0.001\,rad. Sprawling, it gives way at 370 of its 666 hinges, by up to 90 degrees, and comes to rest 46\,cm high, where it was drawn 174\,cm high; after 5\,s in MuJoCo, 596 of its hinges are within 0.05\,rad of the settled pose (median 0.004\,rad), and one truss and one leaf lying across others move further.

\section{Reference traits and scoring}
\label{sec:supp:reference}

\paragraph{Sources.} For each species of the test set we fetched its treatment in the Flora of
China and, where it treats the species, in the Flora of North America, together with the genus and family treatments for what a species treatment leaves to them (leaves alternate, leaves peltate, flowers in a raceme). The reference traits were written from these pages alone, before any run: nothing comes from a model or from our runs. Each value carries its source and a quotation, and a script checks that every quotation appears in the fetched page.

\paragraph{Traits.} The sixteen traits are those that both our growth trace and the organ reader
give, in the states our traits use, so that every system is scored the same way: habit, main stem, side shoots, leaf arrangement and its pattern, leaf type and division, leaflets per leaf, leaflet arrangement, pinnules, leaf shape, margin, lobe arrangement, attachment, inflorescence form and flowers per inflorescence. Where a source allows several states (``narrowly elliptic, elliptic, ovate, or obovate''), all of them agree, and no more than it allows; a count is a range. A compound leaf's shape is its leaflets'. A prompt for a plant in fruit is scored on the flowering or fruiting cluster the flora describes.

\paragraph{What is not scored, and what an absent organ means.} A trait the sources do not
settle is not scored, nor one whose form the flora names but no state of ours does (a pine's fascicled needles and cones, the cladodes of asparagus, the syconium of a fig): no system's asset could be read the same way for it. Where the plant has no such organ at its stage (a simple leaf has no leaflets; a heading cabbage has not flowered), the reference says so, and an asset agrees when it has none either. There are 276 scored traits, 10 to 15 per species; 94 of them say an organ is absent. Any asset lacking the organ meets those, a single rigid body included, so the main text scores the 182 traits of organs the plant has. \Cref{tab:supp:flora} gives all 276.

\paragraph{Scoring.} The organ reader builds the tree of an exported stage's rigid bodies and
joints, classes each body by the head word of its name (stem, leaf, petiole, flower, fruit), and reads shoots, leaves and inflorescences from that tree, a blade's outline and margin from its meshes seen face on. It is the same for every system; the agents' task asks for one rigid body per organ, named after the organ, as our export writes them. A trait agrees when the value read is one the reference allows; a prompt with no asset agrees on none of its traits. For our method the traits are also read from the growth trace, the plant as the description grew it; on its plants on GPT-6-astra, the reader gives the trace's value for 154 of the 181 traits of organs present that either reading gives, and over all 276 traits the trace agrees with the floras on 269 where the reader finds 251. The tables report the reader's values.

\begin{table}[t]
  \centering
  \caption{\textbf{Flora agreement over all 276 scored traits} (GPT-6-astra), including the 94
that say an organ is absent: group A 130, B 70, C 76.}
  \label{tab:supp:flora}
  \footnotesize
  \setlength{\tabcolsep}{4pt}
  \begin{tabular}{@{}lcccc@{}}
    \toprule
    System & A & B & C & All \\
    \midrule
    3D-GPT~\cite{sun2025threedgpt} & 0 & 0 & 0 & 0 \\
    FloraForge~\cite{hadadi2025floraforge} & 6 & 0 & 0 & 6 \\
    Articraft~\cite{articraft} & 43 & 31 & 33 & 107 \\
    Codex & 120 & 53 & 63 & 236 \\
    DeepSeek Harness & 117 & 52 & 63 & 232 \\
    \ours & 123 & 61 & 67 & 251 \\
    \bottomrule
  \end{tabular}
\end{table}

\section{Grounding ablation: further readings}
\label{sec:supp:ablation}

The ablation of \cref{tab:lowknow} is scored with the organ reader that scores every system (\cref{sec:supp:reference}). Here it has a fourth condition, neither search nor profile, and runs on GPT-5.5 as well; the conditions are named by what they keep, so profile only is the main text's w/o search and search only its w/o profiling. Since all four conditions run our method, they can also be read from the growth trace and compared on how their runs went. \Cref{tab:supp:ablation_trace,tab:supp:ablation_process} give both for GPT-6-astra and GPT-5.5.

\paragraph{Growth trace.} The trace gives the plant as its description grew it, before export: its stems and branching, leaf arrangement and division, counts and inflorescence form. Read this way, every condition agrees with the floras on 164 to 175 of the 182 traits, against 143 to 157 by the reader. The one large difference on the trace is search with GPT-5.5 on the less-documented species: 37 of 50 traits with neither search nor profile, 45 and 47 with search. GPT-6-astra reaches 46 to 49 of them without search, so its differences on the reader lie in the form of the organs after export, which only the reader judges.

\begin{figure*}[t]
  \centering
  \includegraphics[width=\linewidth]{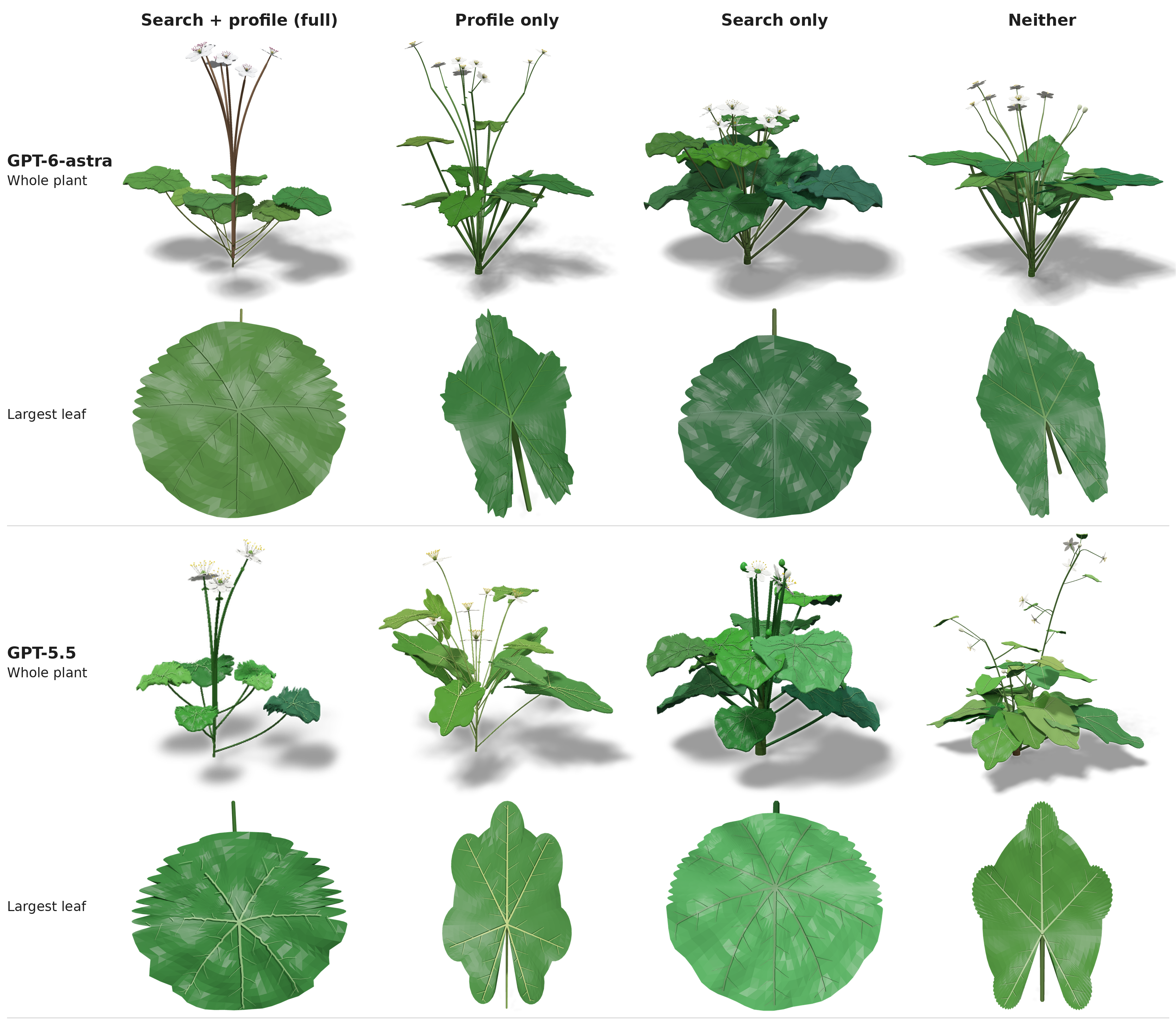}
  \caption{\textbf{\emph{Asteropyrum peltatum} under the four grounding conditions}, with GPT-6-astra (top) and GPT-5.5 (bottom): each final plant (one scale per base model) and its largest leaf seen from above (each leaf at its own scale). The Flora of China describes the leaves as peltate and suborbicular. With search, the blades are round and the petiole joins them near the center; without search, they are cordate with the petiole at the base.}
  \label{fig:supp:ablation_leaf}
\end{figure*}

\paragraph{Organ form.} \Cref{fig:supp:ablation_leaf} shows one such trait. The leaves of \emph{Asteropyrum peltatum} are peltate and suborbicular in the Flora of China. With search, which brings both text and photographs, both base models give it round blades joined by the petiole near their center; without search, with or without a profile, both give a cordate blade with the petiole at its base. Over the six less-documented species, the four conditions disagree on 14 species--trait pairs with GPT-6-astra and 22 with GPT-5.5, mostly leaf shape and attachment and inflorescence form; with GPT-5.5 and neither search nor profile, the habit or main stem of \emph{Beesia}, \emph{Mukdenia} and \emph{Ypsilandra} is wrong as well.

\paragraph{Process.} Grounded runs take longer and build more (\cref{tab:supp:ablation_process}). With GPT-5.5, the median plant has 991 organs with search and profile and 238 with neither, and its authoring takes 22.8 against 5.7 minutes. More organs leave more contacts between them: organ intersections make up most of the final rounds' check failures, which 7 and 6 of 22 plants have with search and profile against 2 and 3 with neither. Without search, the model corrects its profile from its own knowledge alone (7 and 19 corrections over 22 plants); with search, its corrections also cite the photographs and search results.

\begin{table}[t]
  \centering
  \caption{\textbf{The grounding ablation read two ways}: flora agreement over the traits of organs the plant has, from the growth trace and by the organ reader (as in \cref{tab:lowknow}), on the traits of well-documented (132) and less-documented (50) species and on all 182. One profile-only run with GPT-6-astra timed out and agrees on none.}
  \label{tab:supp:ablation_trace}
  \footnotesize
  \setlength{\tabcolsep}{3pt}
  \begin{tabular}{@{}llcccccc@{}}
    \toprule
     & & \multicolumn{3}{c}{Growth trace} & \multicolumn{3}{c}{Organ reader} \\
    \cmidrule(lr){3-5}\cmidrule(lr){6-8}
    Base LLM & Condition & Well & Less & All & Well & Less & All \\
    \midrule
    GPT-6-astra & Neither & 126 & 49 & 175 & 118 & 38 & 156 \\
     & Profile only & 122 & 46 & 168 & 113 & 36 & 149 \\
     & Search only & 126 & 46 & 172 & 112 & 41 & 153 \\
     & Search + profile & 128 & 47 & 175 & 116 & 41 & 157 \\
    \midrule
    GPT-5.5 & Neither & 127 & 37 & 164 & 116 & 27 & 143 \\
     & Profile only & 128 & 43 & 171 & 110 & 37 & 147 \\
     & Search only & 122 & 47 & 169 & 111 & 40 & 151 \\
     & Search + profile & 128 & 45 & 173 & 114 & 37 & 151 \\
    \bottomrule
  \end{tabular}
\end{table}

\begin{table}[t]
  \centering
  \caption{\textbf{How the ablation's runs went}, over the 22 test prompts: plants built; plants whose final round fails a check; the median number of organs (rigid bodies) of the final plant; the mean number of rounds; the median authoring time (wall clock, with the four conditions run at the same time); the mean number of search calls (queries and page opens) and of reference photographs per plant; and the profile corrections made during revision, in all.}
  \label{tab:supp:ablation_process}
  \footnotesize
  \setlength{\tabcolsep}{2.5pt}
  \resizebox{\linewidth}{!}{%
  \begin{tabular}{@{}llccccccc@{}}
    \toprule
    Base LLM & Condition & Built & \shortstack{Check\\failures} & Organs & Rounds & \shortstack{Authoring\\(min)} & \shortstack{Search calls /\\photos} & \shortstack{Profile\\corrections} \\
    \midrule
    GPT-6-astra & Neither & 22 & 2 & 535 & 2.5 & 3.0 & -- & -- \\
     & Profile only & 21 & 4 & 741 & 3.3 & 6.8 & -- & 7 \\
     & Search only & 22 & 4 & 559 & 2.9 & 7.1 & 6.8 / 3.3 & -- \\
     & Search + profile & 22 & 7 & 619 & 3.5 & 8.4 & 5.4 / 3.2 & 6 \\
    \midrule
    GPT-5.5 & Neither & 22 & 3 & 238 & 3.2 & 5.7 & -- & -- \\
     & Profile only & 22 & 3 & 683 & 3.7 & 13.3 & -- & 19 \\
     & Search only & 22 & 3 & 615 & 3.5 & 16.2 & 4.7 / 4.5 & -- \\
     & Search + profile & 22 & 6 & 991 & 3.8 & 22.8 & 6.4 / 4.4 & 7 \\
    \bottomrule
  \end{tabular}}
\end{table}

\section{Further results across base LLMs}
\label{sec:supp:models}

\paragraph{A fourth base LLM.}
\Cref{tab:supp:gemma} repeats the comparison of \cref{tab:compare} with Gemma~4~31B, served on our
own GPUs like Qwen3.8-27B. 3D-GPT runs on GPT-6-astra only.

\begin{table}[t]
  \centering
  \caption{\textbf{Quantitative comparison with Gemma~4~31B}, as in \cref{tab:compare}: prompts with a generated asset (22), reference traits met on well-documented (132), less-documented (50) and all (182) species, and numerical constraints met on the detailed prompts (66). Best scores in bold (ties included).}
  \label{tab:supp:gemma}
  \footnotesize
  \setlength{\tabcolsep}{2pt}
  \begin{tabular}{@{}llccccc@{}}
    \toprule
    Category & Method & Generated & Well & Less & All & Constraints \\
    \midrule
    Tailor-made & Codex & 19 & 41 & 20 & 61 & 28 \\
     & DeepSeek Harness & \textbf{21} & 40 & \textbf{28} & 68 & 31 \\
     & Articraft~\cite{articraft} & 19 & 6 & 7 & 13 & 18 \\
    \midrule
    Mass-grown & FloraForge~\cite{hadadi2025floraforge} & 1 & 0 & 0 & 0 & 7 \\
     & \ours & 20 & \textbf{91} & \textbf{28} & \textbf{119} & \textbf{43} \\
    \bottomrule
  \end{tabular}
\end{table}

\paragraph{Population consistency on open-weight base LLMs.}
Profile consistency depends on whether authoring ends with a description that passes the profile
checks. With Qwen3.8-27B, 11 of the 21 final descriptions pass every profile check in their last
round, and 95.8\% of their unseen individuals agree with the profile, close to GPT-6-astra
(98.7\% over its 21 such descriptions) and GPT-5.5 (99.4\% over 22). The other 10 still disagree
with the profile when the revision budget is spent, and 5.1\% of their unseen individuals agree.
Gemma~4~31B splits the same way: 6 descriptions pass (86.3\%) and 14 do not (3.9\%), for an overall
consistency of 30.2\% during authoring and 28.6\% on unseen seeds. For every base LLM the two seed
sets differ by less than two percentage points, so the lower consistency comes from descriptions
the agent did not bring into agreement with the profile within the revision budget, not from
variation across individuals. The traits left in disagreement are spread over many organs,
including bracts, calyx, petal markings, leaf apex, fruit shape and stem counts.

\paragraph{Reasoning effort.}
Nine test prompts, three from each group, were also run at medium and high reasoning effort on
GPT-6-astra and GPT-5.5 by Codex, DeepSeek Harness, Articraft and ours. \Cref{tab:supp:effort} counts flora
agreement on the 73 reference traits of organs the plant has. At low effort our method agrees on 62
and 60 traits, more than any other system at any effort (at most 60 on GPT-6-astra and 56 on
GPT-5.5). Higher effort changes a few of our traits on two to four species in either direction (59
and 59 on GPT-6-astra, 57 and 60 on GPT-5.5). Codex and DeepSeek Harness gain with effort on
GPT-6-astra (55 to 60 and 53 to 57), DeepSeek Harness also on GPT-5.5 (40 to 56), and Articraft
stays at 7 to 8.

\begin{table}[t]
  \centering
  \caption{\textbf{Reasoning effort}: reference traits met on nine test prompts (73 traits of organs present) at low, medium and high effort. Best scores in bold.}
  \label{tab:supp:effort}
  \footnotesize
  \setlength{\tabcolsep}{3pt}
  \begin{tabular}{@{}lcccccc@{}}
    \toprule
     & \multicolumn{3}{c}{GPT-6-astra} & \multicolumn{3}{c}{GPT-5.5} \\
    \cmidrule(lr){2-4}\cmidrule(l){5-7}
    Method & Low & Medium & High & Low & Medium & High \\
    \midrule
    Codex & 55 & 56 & \textbf{60} & 38 & 50 & 44 \\
    DeepSeek Harness & 53 & 54 & 57 & 40 & 45 & 56 \\
    Articraft~\cite{articraft} & 7 & 8 & 8 & 8 & 8 & 8 \\
    \ours & \textbf{62} & \textbf{59} & 59 & \textbf{60} & \textbf{57} & \textbf{60} \\
    \bottomrule
  \end{tabular}
\end{table}

\paragraph{Revision budget.}
The coding agents have no limit on revisions or tool calls, only on time: four hours a prompt (Codex
six on Qwen3.8-27B). Our method writes a first description and at most three revisions, within three
hours. \Cref{tab:supp:budget} shows what each system used on the species prompts. On GPT-6-astra and
GPT-5.5, both agents end every run on their own, after a median of 1.4 to 3.3 minutes and 5 to 11
shell commands, with most of their time unused; our method takes a median of 12 and 15 minutes and
uses every revision on 13 and 19 of the 22 prompts. At high reasoning effort the agents stay below
our method at low effort (\cref{tab:supp:effort}). On Qwen3.8-27B the agents reach their time limit
on 6 (Codex) and 18 (DeepSeek Harness) of the 22 prompts, and our method on one.

\begin{table}[t]
  \centering
  \caption{\textbf{Budget used} on the 22 species prompts: median wall time per prompt (Minutes), runs stopped at their time limit (Limit: four hours for the coding agents, six for Codex on Qwen3.8-27B, three for ours), and median shell commands of the coding agents (Commands) or runs of ours that used every revision (Revisions).}
  \label{tab:supp:budget}
  \footnotesize
  \setlength{\tabcolsep}{2pt}
  \begin{tabular}{@{}llrrrr@{}}
    \toprule
    Base LLM & Method & Minutes & Limit & Commands & Revisions \\
    \midrule
    GPT-6-astra & Codex & 3.3 & 0 & 10.5 & -- \\
     & DeepSeek Harness & 1.4 & 0 & 5 & -- \\
     & \ours & 12.0 & 0 & -- & 13 \\
    \midrule
    GPT-5.5 & Codex & 1.9 & 0 & 11 & -- \\
     & DeepSeek Harness & 1.5 & 0 & 6.5 & -- \\
     & \ours & 15.3 & 0 & -- & 19 \\
    \midrule
    Qwen3.8-27B & Codex & 267.5 & 6 & 241 & -- \\
     & DeepSeek Harness & 240.0 & 18 & 72.5 & -- \\
     & \ours & 65.3 & 1 & -- & 21 \\
    \bottomrule
  \end{tabular}
\end{table}

\section{Physical simulation}
\label{sec:supp:physics}

We simulate the plant our method made for each species prompt of the test set on each base LLM, as exported: 85 plants, of which 22, 22, 21 and 20 on GPT-6-astra, GPT-5.5, Qwen3.8-27B and Gemma~4~31B. The settings are under \emph{Simulation} in \cref{sec:supp:impl}, and \cref{fig:supp:push} shows one push and one pick.

\paragraph{MuJoCo.}
All 85 plants load in MuJoCo~\cite{todorov2012mujoco} and every test completes. The push compares what a load does to the plant with predictions made from the file (\cref{tab:supp:mujoco}). Without gravity, the main stem moves within 1\% of what the file's hinge chain predicts in every plant, and within 5\% of the continuous beam's prediction in 81. With gravity and the springs' preload, it moves within 5\% of the prediction from $\mathbf K_{\mathrm{eff}}$ in 77 of the 83 plants that the stability check passed; three of the other six fold when nudged. Nudged at every hinge, 72 plants return to the drawn pose, 7 are still returning after 15\,s, and 6 fold.

\begin{table*}[t]
  \centering
  \caption{\textbf{Our test-set plants in MuJoCo}, one per species prompt and base LLM. Push without gravity: the main stem's movement over the prediction of the file's hinge chain (Chain) and of the continuous tapered beam (Beam), median and range. Push with gravity (Loaded): plants within 5\% of the prediction from $\mathbf K_{\mathrm{eff}}$, of those the stability check passed. Nudge: plants that return to the drawn pose, are still returning after 15\,s, or fold.}
  \label{tab:supp:mujoco}
  \footnotesize
  \setlength{\tabcolsep}{6pt}
  \begin{tabular}{@{}lrcccrrr@{}}
    \toprule
     & & \multicolumn{3}{c}{Push} & \multicolumn{3}{c}{Nudge} \\
    \cmidrule(lr){3-5}\cmidrule(l){6-8}
    Base LLM & Plants & Chain & Beam & Loaded & Return & Returning & Fold \\
    \midrule
    GPT-6-astra & 22 & 1.000 [1.000, 1.001] & 1.001 [0.961, 1.012] & 21/22 & 20 & 1 & 1 \\
    GPT-5.5 & 22 & 1.000 [1.000, 1.001] & 0.999 [0.836, 1.024] & 20/22 & 20 & 1 & 1 \\
    Qwen3.8-27B & 21 & 1.000 [0.999, 1.009] & 1.007 [0.986, 1.082] & 18/20 & 18 & 0 & 3 \\
    Gemma~4~31B & 20 & 1.000 [0.993, 1.001] & 1.007 [0.815, 1.125] & 18/19 & 14 & 5 & 1 \\
    \bottomrule
  \end{tabular}
\end{table*}

\paragraph{Isaac Sim.}
The 22 plants made on GPT-6-astra are held for 3\,s from the drawn pose. A plant \emph{stands} if every body's pose stays finite, no joint breaks and no body moves more than 2\,cm. PhysX limits an articulation to 255 links and does not break joints inside one, so a larger plant keeps the rest of its hinges, and every plant its fruit joints, as ordinary joints, which do not hold chains of light bodies at our settings. As exported, 9 plants stand (\cref{tab:supp:isaac}): 5 of the 6 that fit one articulation and 4 of the 16 that do not. At PhysX's default solver setting, 7 stand. With the bodies regrouped at load time into articulations of at most 255 links and the file unchanged, 14 stand, at 0.7 to 17 times the cost of a step. We then pull five fruit joints of each fruiting plant, 51 joints on 11 plants (the pineapple has a single fruit). On the five fruiting plants that stand once regrouped, 22 of the 25 joints break within 2\% of the force the model chose, and three sorghum grains separate earlier, at 0.16 to 0.19 of it. On the plants that do not stand, the fruit come off within the first steps, at no more than 0.08 of the break force, except the pineapple's, which comes off at that force.

\begin{table}[t]
  \centering
  \caption{\textbf{Our GPT-6-astra plants in Isaac Sim.} Plants of the 22 that stand through a 3\,s hold (Stand), among the 6 of at most 255 bodies and the 16 larger ones; the largest movement of a body in a standing plant (Moved, in mm); and the fruit joints pulled on standing plants that break within 2\% of their break force (At force).}
  \label{tab:supp:isaac}
  \footnotesize
  \setlength{\tabcolsep}{3pt}
  \begin{tabular}{@{}lccccc@{}}
    \toprule
    Setting & Stand & $\leq$255 & $>$255 & Moved & At force \\
    \midrule
    As exported, PGS 1000\,Hz & 9 & 5 & 4 & 3.5 & 6/10 \\
    As exported, TGS 60\,Hz & 7 & 4 & 3 & 10.1 & -- \\
    Regrouped, PGS 1000\,Hz & 14 & 5 & 9 & 10.2 & 22/25 \\
    \bottomrule
  \end{tabular}
\end{table}

\begin{figure*}[t]
  \centering
  \footnotesize
  \setlength{\tabcolsep}{0pt}
  \newcommand{\cell}[1]{\raisebox{-.5\height}{\includegraphics[width=0.128\linewidth]{figs/physics/#1}}}
  \begin{tabular}{@{}p{0.09\linewidth}*{7}{c}@{}}
    \textbf{Push}\newline MuJoCo & \cell{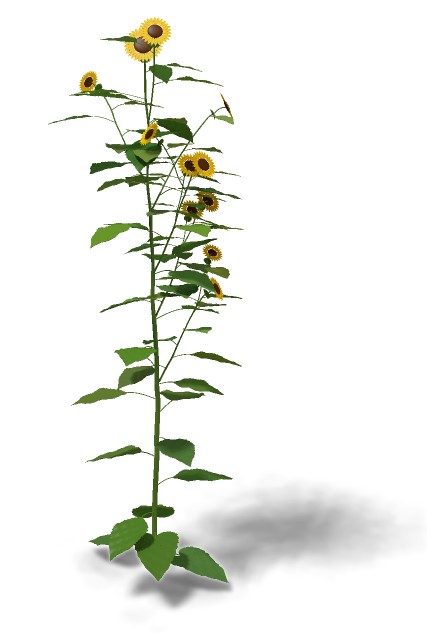} & \cell{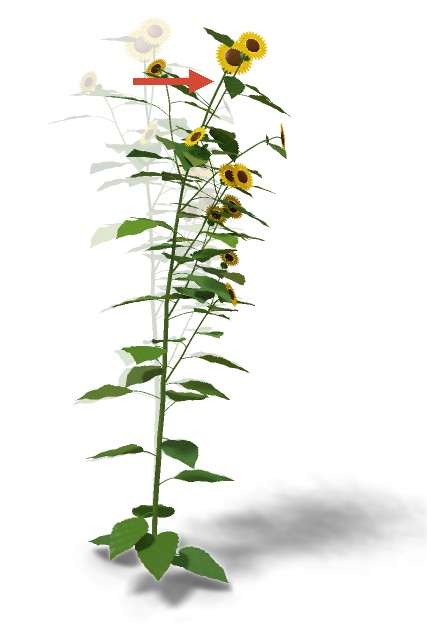} & \cell{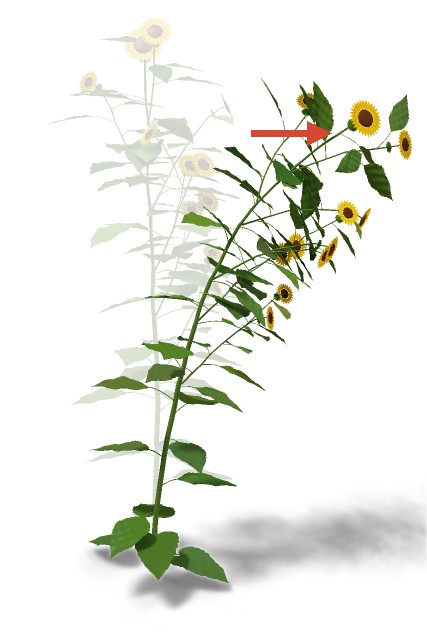} & \cell{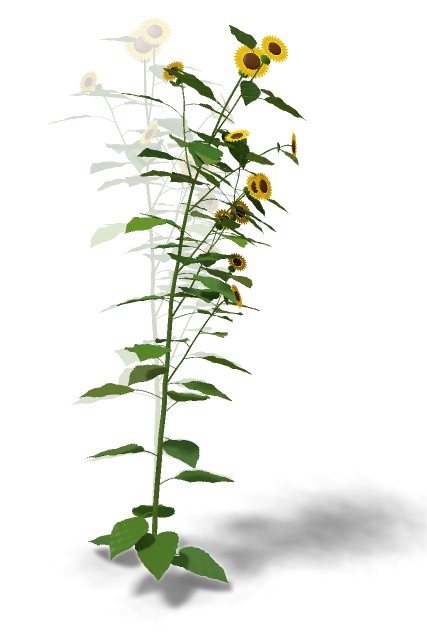} & \cell{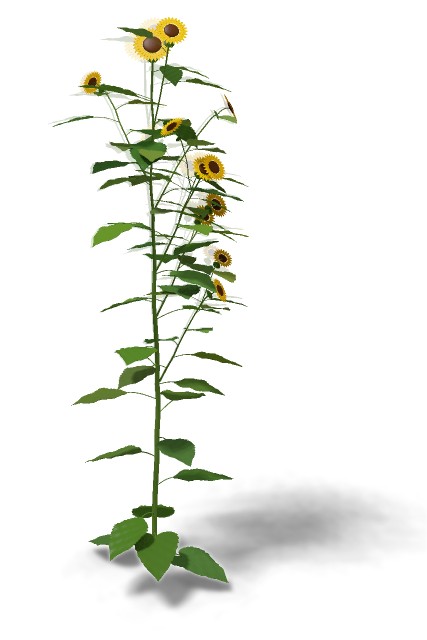} & \cell{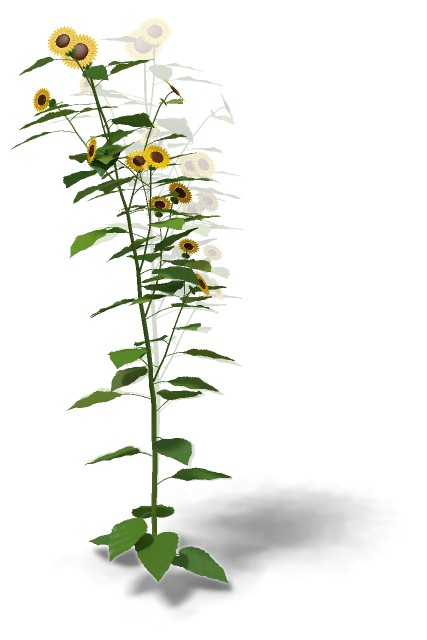} & \cell{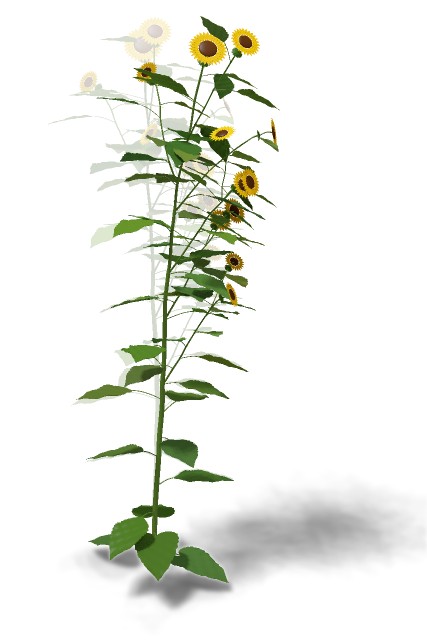} \\
    & 0\,s & 1\,s & 2\,s & 3\,s, released & 3.5\,s & 3.9\,s & 6\,s \\[4pt]
    \textbf{Pick}\newline Isaac Sim & \cell{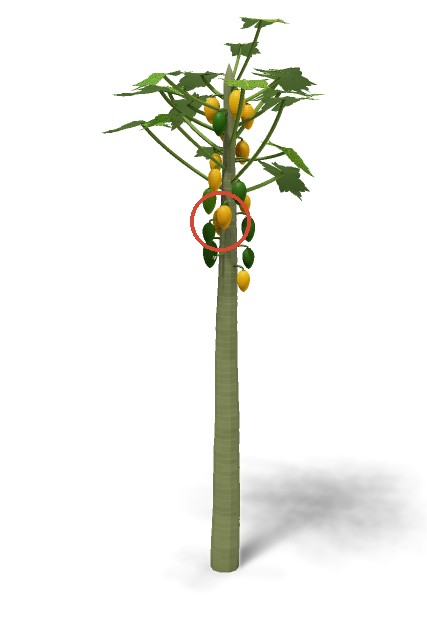} & \cell{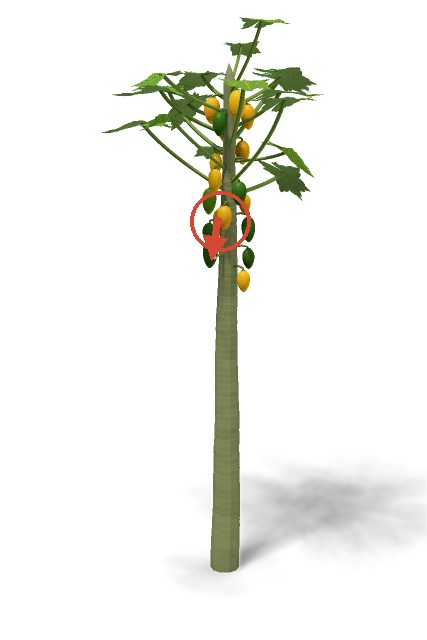} & \cell{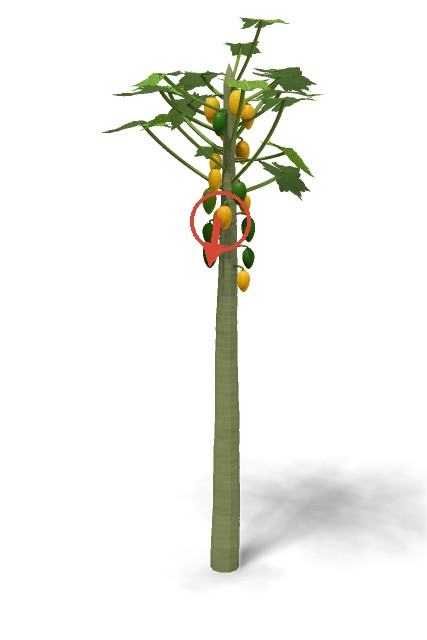} & \cell{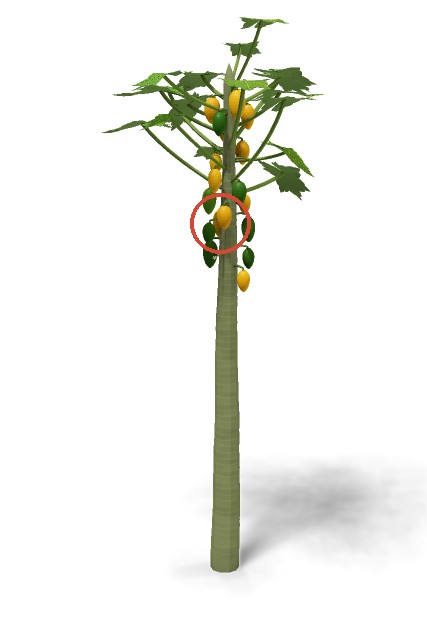} & \cell{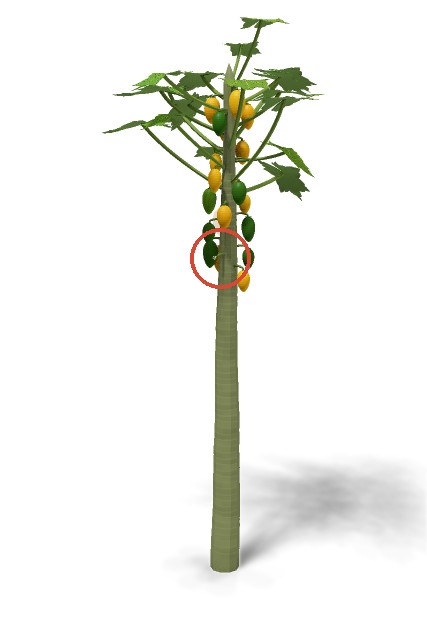} & \cell{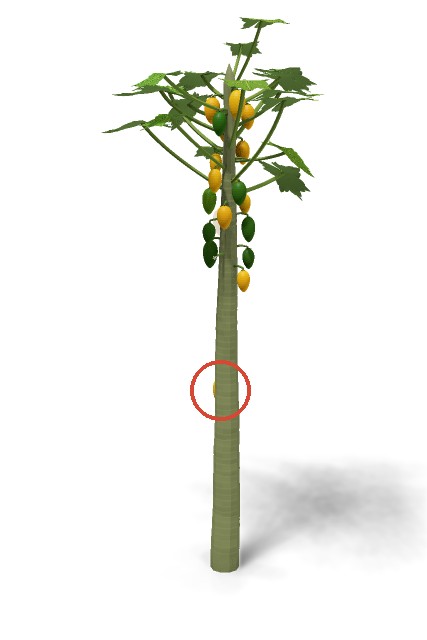} & \cell{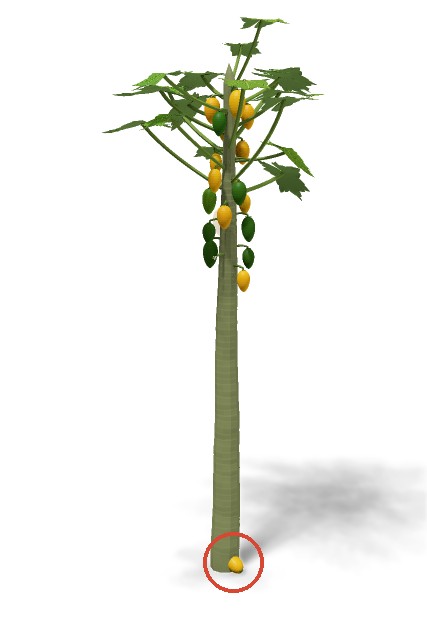} \\
    & 0\,s & 0.4\,s & 0.6\,s & 0.8\,s, off & 1.0\,s & 1.3\,s & 2\,s \\
  \end{tabular}
  \caption{\textbf{Our plants in simulation.} Top: the sunflower of the human-study set, pushed at
    the top of its main stem in MuJoCo; the load rises, is held and is released, and the stem
    bends at its hinges and swings back. Bottom: a papaya of the test set in Isaac Sim, one fruit
    (ringed) pulled along its stalk until its joint breaks at the force the model chose; the fruit
    falls and the others stay on. The drawn pose is faint behind the later frames, and arrows mark
    the loads while they act.}
  \label{fig:supp:push}
\end{figure*}

\paragraph{Supplementary video.}
The video opens on a garden grown from nine descriptions, 30 individuals of each, every bed labeled with its prompt: seven descriptions of the human-study set, an eggplant authored with the same settings, and the test set's papaya, all on GPT-6-astra. A wind then crosses the 270 plants in MuJoCo, at 4.5\,m\,s$^{-1}$ on average with gusts, and three fruit of the papaya are picked in Isaac Sim as in \cref{fig:supp:push}. The last part turns the assets of \cref{fig:comparison}, for its three prompts and both base LLMs. For \emph{Beesia calthifolia} and \emph{Mukdenia rossii}, each system's asset is then loaded in Isaac Sim as that system left it, on a floor, under gravity and then one gust of wind, the same for every asset, at 6\,m\,s$^{-1}$ on average. Our four plants sway by 4 to 21\,cm and return to within 0.5\,cm of the drawn pose. Of the twelve assets of the other systems, seven move by at most 5\,mm, two of them because the simulator holds none or only one of their bodies; one sways and returns; and four sag, come apart or are carried off. The rosemary is not simulated: our rosemary on GPT-6-astra, with 10{,}150 bodies, does not stand in Isaac Sim.

\section{Further limitations}
\label{sec:supp:limits}

\paragraph{Evaluation.}
The reference traits come from flora treatments that the model is not shown, but a model that searches may read the same treatments. Flora agreement with search therefore measures how far the method carries what the sources say into the plant, as well as what the model knows. The organ reader that scores every system misreads some plants of every system; on our plants of the test set on GPT-6-astra it gives the growth trace's value for 154 of the 181 traits of organs present that either reading gives.

\paragraph{Validation against real plants.}
The moduli, densities and detachment forces are chosen by the model, from sources where it finds them. That a simulated stem bends as beam theory predicts shows that the simulator reproduces the file, not that the file is right. Likewise, the individuals of one description vary as the model's ranges say; we have not compared this variation with that of measured populations, nor the morphology of generated plants (internode lengths, branching angles, leaf area) with that of measured plants. The wind in the supplementary video is a drag rule applied to each rigid body; no flow of air is simulated.

\paragraph{Vocabulary coverage.}
The vocabulary covers the organs that the floras of the plants we developed it on describe. For a plant with organs it lacks, such as needles and cones, catkins, succulent stems or underground organs, the model describes the plant as near as the vocabulary allows and reports what it could not draw; group B of the test set measures how far this goes (\cref{tab:supp:flora}). The pineapple of that group is one such plant: its fruit, fused from the fruits of a whole spike, and the crown of leaves above it are outside the vocabulary, and the plant is drawn with a single fruit body and no crown.

\paragraph{Supports and rigid organs.}
A staked or twining plant's main stem is held rigid rather than tied, so it does not bend between ties, and its stake or pole carries no load. A sprawling plant settles without contact between its organs and without friction. Leaves are rigid bodies hinged at their bases, without the lamina's own give.

\paragraph{Simulation.}
In Isaac Sim, eight of the 22 plants do not stand even with their bodies regrouped (\cref{tab:supp:isaac}). Two of them have no fruit joints, and we have not established why they fail; the pineapple, which fits one articulation, drops its one fruit during the hold. The same files load in MuJoCo, where all are simulated. There, six of the 85 plants fold when nudged, four of which the stability check had passed: the check takes its margin in the direction of least effective stiffness, usually that of a soft hinge that carries little load, so a plant can pass while its weight has nearly used up the stiffness of its main stem.

\section{Implementation details}
\label{sec:supp:impl}

\paragraph{Settings.}
On the test set every system runs at low reasoning effort, and our method searches through the self-hosted search engine on every model. The twenty-four plants of the human study were made by every system on GPT-6-astra at high reasoning effort, the other systems' before the test set. Our method's plants were made with the version evaluated on the test set, searching through the same search engine with three attempts at each call.

\paragraph{Models and software.}
GPT-6-astra and GPT-5.5 are called through OpenAI's API as \texttt{gpt-6-astra} and
\texttt{gpt-5.5}. Qwen3.8-27B and Gemma~4~31B (\texttt{gemma-4-31B-it}) are served on our own GPUs
by vLLM~0.32.0, each on two GPUs with a 131{,}072-token context. All test-set runs took place
between 11 and 17 November 2026. Codex runs as Codex CLI~0.167.1 (0.169.0 for the detailed prompts
on GPT-5.5, Qwen3.8-27B and Gemma~4~31B), and DeepSeek Harness as version~0.2.1-rc.5.

\paragraph{Agent interface and control.}
A single LLM agent performs botanical research, profile construction, developmental description authoring, visual review, and revision. The harness exposes operations for web search and page retrieval, reference-image selection, and recording or editing the profile and description. Recording a profile triggers schema and cross-entry checks. Recording or editing a description triggers validation, growth, compilation, and checks, whose results are returned with measurements and rendered views. The LLM interprets this feedback and decides whether to accept the plant or revise the description; the engine, compiler, and checkers execute the corresponding operations programmatically.

\paragraph{Full vocabulary.}
\label{sec:supp:vocab}
The description's schema names 313 fields. Each description is a typed record emitted through a tool call and checked against the schema before execution. It contains shoot archetypes and their references, an initial archetype, count and height targets, and support conditions. Botanical choices have no implicit defaults: organ forms and physical traits must be selected explicitly. Values are expressed as constants, ranges, log-normal distributions, Bernoulli probabilities, or categorical choices as appropriate to the field. Neither the growth engine nor the compiler selects a species-specific template. Each archetype gives:
\begin{itemize}[nosep,leftmargin=*]
\item \emph{Axis and stem}: orthotropic or plagiotropic; the number of phytomers it may grow; internode lengths and how they change along the shoot; the stem's radius, lean, waviness and cross-section (round, angled, ribbed or winged); the angle its branches leave at; prickles, thorns or bulbils; the share of its lower nodes whose leaves have fallen.
\item \emph{Phyllotaxis}: leaves per node and the divergence angle between successive nodes, with an individual jitter.
\item \emph{Leaves}: simple or compound, divided once, twice or three times (pinnate, palmate or ternate at each depth), with leaflet and pinnule counts; the blade's outline, apex and base (oblique, auricled, clasping, perfoliate or decurrent), its lobes and their arrangement, teeth and their shape, venation and markings, a peltate attachment; droop, cupping, undulation, blistering, texture and the colors of both faces; the petiole, stipules, a sheath with its ligule, an ochrea; tendrils at the leaf's tip; juvenile leaves on the first nodes.
\item \emph{Meristem fates}: at each node, the chances that the apex turns reproductive or dormant and how they change along the shoot; the same for axillary buds, the band of nodes whose buds may grow, and the archetype a growing bud becomes; forks that split an apex into two or more shoots; determinate growth.
\item \emph{Inflorescences}: the form; flowers per inflorescence and per node; posture (erect, arching or nodding), flowers on one side, a coiled tip; bracts below each flower, in pairs, at forks or around the base; sterile marginal flowers; male flowers, where they stand and their own form; whether flowers that set no fruit stay open or are shed, and how many are in bud.
\item \emph{Flowers}: the corolla (separate petals, fused, bell-shaped, tubular, two-lipped, papilionaceous, a composite head with rays and disk, or none) with its span, posture, doubling and spurs; petals and sepals that differ by where they stand, as a flora describes an irregular flower's upper, lateral and lower petals; a base blotch, spots, an outer color, a corona; stamens and their posture, a staminal column, staminodes, carpels with style and stigma, an inferior ovary, a column fusing stamens and style; whether the flower nods.
\item \emph{Fruit}: size, shape, surface and color; whether it hangs or is held, and at what angle; how far each cluster has ripened; awns, hulls, cupules and wings; heads of achenes or of plumed seeds; whether the fruit splits open, and what it shows.
\item \emph{Physical traits}: the Young's modulus of the stems and of the fruit stalks (MPa), the stems' ratio of twisting to bending compliance, the fruit's density (kg\,m$^{-3}$) and the force that pulls a ripe fruit off (N).
\end{itemize}
Targets are ranges for the number of side shoots, main branches, inflorescences, fruit, ripe fruit and open flowers, and for the plant's height.

\paragraph{Complete description example.}
\label{sec:supp:description_example}
The listing below is the final description our method wrote on GPT-6-astra for the test prompt
``a flowering \emph{Asteropyrum peltatum} plant''; its individual grown from seed 0 is the plant at
the top left of \cref{fig:supp:ablation_leaf}. It is printed in full; only line breaks and spaces
differ from the file. The plant has one archetype, a crown of peltate, orbicular, crenate leaves
whose axillary buds bear solitary flowers with five white sepals and eight yellow, spatulate petals.
A \texttt{null} marks a field the plant does not use, and the prompt states no targets. Lengths are
in meters except proportions such as \texttt{stamen\_length}, a share of the flower's size; angles
are in radians, and moduli, density and detachment force are in the units given above.
All of individuals 1 to 100 of this description agree with its profile.

{\scriptsize
\begin{verbatim}
{
 "root_archetype": "crown",
 "archetypes": [
  {
   "id": "crown", "axis_orientation": "orthotropic",
   "expected_span": 7, "max_phytomers": 9,
   "radius_start": 0.00085, "radius_end": 0.0007,
   "stem": {
    "color": [0.3, 0.32, 0.16, 1.0], "base_tilt": 0.1,
    "start_elevation": 1.5, "bend_toward_tip": 1.2,
    "waviness": 0.07, "modulus": 700,
    "flexibility": 1.0, "twist_to_bend": 3,
    "section": null, "section_count": null,
    "section_depth": null, "prickles": null,
    "thorns": null, "bulbils": null
   },
   "phyllotaxis": {
    "divergence_angle": 2.4, "leaves_per_node": 1,
    "divergence_jitter": 0.2
   },
   "internode_length_base": {
    "type": "uniform", "low": 0.001, "high": 0.003
   },
   "internode_length_profile": "uniform",
   "fate_pattern": "uniform",
   "reproductive_probability": 0,
   "dormant_probability": 0,
   "axillary_fate_pattern": "uniform",
   "axillary_reproductive_probability": 0.45,
   "axillary_dormant_probability": 0.55,
   "child_archetype": null, "vigor_decay": 0.75,
   "determinate": true,
   "leaf": {
    "size": {
     "type": "uniform", "low": 0.025, "high": 0.06
    },
    "width_ratio": 1.05, "compound": false,
    "leaflet_count": 1.0, "compounding_depth": 1,
    "pinnule_count": null, "margin_serration": 0.08,
    "serration_from": 0.0,
    "color": [0.22, 0.38, 0.18, 1.0],
    "petiole_color": [0.38, 0.42, 0.22, 1.0],
    "attachment_tilt": {
     "type": "uniform", "low": -0.18, "high": 0.22
    },
    "leaflet_splay": 1.3,
    "leaflet_arrangement": "palmate",
    "petiole_length": {
     "type": "uniform", "low": 0.03, "high": 0.1
    },
    "petiole_elevation": {
     "type": "uniform", "low": 0.45, "high": 1.1
    },
    "shape": "orbicular", "apex": "rounded",
    "lobe_count": 0, "lobe_depth": 0.4,
    "lobe_arrangement": "palmate", "lobe_shape": null,
    "lobe_tip_lobes": 0, "lobe_tip_depth": 0.3,
    "lobe_width": null, "outer_lobe_length": null,
    "peltation": 0.45, "droop": 0.08,
    "rachis_droop": null, "cupping": 0.03,
    "undulation": 0.025, "undulation_count": 7,
    "bullation": 0.035, "color_variation": 0.045,
    "young_color": [0.34, 0.48, 0.24, 1.0],
    "underside_color": [0.38, 0.49, 0.3, 1.0],
    "size_profile": {
     "shape": "basipetal", "amount": 0.2
    },
    "sheath": null, "ligule": null,
    "sheath_auricles": null, "leaflet_taper": null,
    "terminal_leaflet": null, "petiolule": null,
    "teeth": 18, "tooth_irregularity": 0.3,
    "doubly_serrate": false, "tooth_shape": "crenate",
    "apex_mucro": null, "cordate_tip": null,
    "lobe_tip": null, "lobes_to": null,
    "pattern": null, "pattern_color": null,
    "pattern_extent": null, "pattern_length": null,
    "pattern_feather": null, "pattern_veins": null,
    "venation": "palmate", "vein_count": 7,
    "vein_color": [0.27, 0.42, 0.21, 1.0],
    "vein_network": true, "thickness": null,
    "base_oblique": null, "base_auricles": null,
    "sessile_base": null, "decurrent_length": null,
    "tip_tendrils": null, "stipules": null
   },
   "fruit": {
    "diameter": 0.006, "aspect_ratio": 1.3,
    "color": [0.4, 0.29, 0.15, 1.0],
    "unripe_color": [0.29, 0.48, 0.2, 1.0],
    "ripening": {
     "min": 0.0, "max": 0.0, "front": 0.2,
     "irregularity": 0.0
    },
    "taper": 0.2, "color_jitter": 0.03,
    "flower_count": 1.0, "fruit_set_probability": 0,
    "flower": {
     "size": {
      "type": "uniform", "low": 0.015, "high": 0.025
     },
     "color": [0.98, 0.81, 0.08, 1.0],
     "petal_style": "separate_petals",
     "petal_count": 8,
     "stamen_color": [0.56, 0.26, 0.4, 1.0],
     "center_fraction": 0.2,
     "sepals": {
      "count": 5, "span": 1, "width_ratio": 0.75,
      "color": [0.97, 0.98, 1.0, 1.0],
      "posture": "spreading", "shape": "obovate",
      "apex": "rounded", "cupping": 0.04,
      "recurve": 0.03, "rise": 0.08,
      "base_color": null, "base_share": null,
      "undulation": 0.025, "tube": null,
      "spotted": false, "places": null
     },
     "corolla_span": 0.25, "petal_posture": "erect",
     "nodding": false, "petal_shape": "spatulate",
     "petal_width_ratio": 0.35,
     "petal_apex": "rounded", "pedicel_length": 0.008,
     "pedicel_posture": null, "petal_tip_lobes": 0,
     "petal_tip_depth": 0.3, "petal_cupping": 0.3,
     "petal_recurve": 0, "bell_shape": null,
     "bell_lobe_depth": null, "floral_tube": null,
     "stamen_length": 0.27,
     "stamen_posture": "spreading", "petal_rise": 1.2,
     "stamen_count": 22, "staminode_count": 0,
     "carpel_count": 5, "staminode_color": null,
     "carpel_color": [0.34, 0.52, 0.22, 1.0],
     "style_length": 0.08, "style_posture": "straight",
     "stigma": "capitate", "stigma_lobes": null,
     "style_color": [0.86, 0.89, 0.69, 1.0],
     "petal_base_color": null,
     "petal_base_share": null, "petal_undulation": 0,
     "ovary_size": 0.05, "petal_spots": null,
     "petal_spot_color": null,
     "petal_spot_shape": null, "petal_whorls": 1,
     "spurs": null, "spur_length": null,
     "spur_curve": null, "corona": null,
     "corona_length": null, "corona_color": null,
     "disk_shape": null, "disk_height": null,
     "upper_lip_lobes": null, "lower_lip_lobes": null,
     "lower_lip_middle": null, "lower_lip_notch": null,
     "upper_lip_arch": null, "inferior_ovary": false,
     "androgynophore": null, "corona_bands": null,
     "tube_length": null, "tube_width": null,
     "tube_curve": null, "bell_length": null,
     "star_depth": null,
     "filament_color": [0.95, 0.96, 0.9, 1.0],
     "outer_color": null, "inner_whorl_color": null,
     "petal_places": null, "spur_side": null,
     "staminal_column": null, "column": null,
     "column_color": null
    },
    "calyx_color": [0.37, 0.46, 0.24, 1.0],
    "stalk_color": [0.4, 0.3, 0.22, 1.0],
    "stalk_modulus": 900, "stalk_flexibility": 1.0,
    "truss_declination": -1.35,
    "peduncle_length": 0.15,
    "peduncle_elevation": 1.48,
    "peduncle_diameter": 0.002,
    "fruiting_pedicel_length": null, "shape": "round",
    "lobe_count": 4, "lobe_depth": 0.2,
    "curvature": 0.0, "inflorescence_form": "solitary",
    "density": 400, "detachment_force": 0.15,
    "detachment": null, "dehiscence": null,
    "flowers_per_node": 1, "umbel_shape": "dome",
    "unset_flowers": "open", "orientation": "erect",
    "awn": null, "rays": null, "rachis_length": null,
    "cyme_width": null, "rachis_turn": null,
    "rachis_coil": null, "secund": false,
    "distichous": false, "panicle_branches": null,
    "panicle_branch_angle": null,
    "panicle_branch_length": null, "bud_share": 0.1,
    "bud_size": 0.6, "bud_aspect": 1.4,
    "drupelets": null, "cupule": null,
    "cupule_scales": false, "furrow": null,
    "achenes": null, "achene_spread": null,
    "surface": null, "surface_size": null,
    "surface_count": null, "surface_width": null,
    "surface_color": null, "wings": null,
    "wing_width": null, "wing_largest": null,
    "bud_color": [0.83, 0.86, 0.76, 1.0],
    "bud_nodding": false, "stigmatic_disc": null,
    "disc_rays": null, "disc_color": null,
    "apical_pores": null, "silks": null,
    "silk_color": null, "sterile_flowers": null,
    "sterile_size": null, "sterile_petals": null,
    "sterile_color": null, "male": null, "bracts": null
   },
   "fork": null, "tendril": null,
   "juvenile_leaf": null, "juvenile_nodes": 0,
   "bare_below": 0, "axillary_zone": null,
   "juvenile_leaves_per_node": 1
  }
 ],
 "targets": {
  "branches": null, "main_branches": null,
  "inflorescences": null, "fruit": null,
  "flowers": null, "ripe_fruit": null, "height": null
 },
 "support": "self_supporting", "twining": null
}
\end{verbatim}
}

\paragraph{The botanical profile.}
A profile records 81 traits, 14 measurements and 12 short descriptions. A trait the plant lacks at the stage asked for is recorded as absent. When a profile is recorded, its entries are validated one by one (states, ranges and the source each cites) and then against each other (for instance a corolla with no inflorescence, alternate leaves in decussate pairs); every problem is returned at once and corrected entry by entry, not rewritten.

Each entry records its basis: the prompt, a cited web source, a reference photograph, the LLM's own knowledge, or inference from a named relative. Explicit source statements determine the corresponding traits, and photographs help resolve visual characteristics the text leaves open. A later correction to the profile is permitted when it misrepresents the requested plant, and its reason is recorded.

\paragraph{Search and photographs.}
The self-hosted SearXNG~\cite{searxng} metasearch instance queries Google's programmable search, Yahoo and Wikipedia. The model has two functions: one returns the top eight results of a query with their snippets, the other up to 8,000 characters of a page; a step may make eight such calls. Every query and page is cached with its date, so that a run can be replayed from the cache alone. Photographs are looked up on Wikimedia Commons and iNaturalist; the model sees the candidates for each view and chooses at most one living, openly licensed plant per view, or none.

The views are the whole plant, a leaf, a flower, an inflorescence, and a fruit. These photographs are presented as references for the species and stage, not as observations of the particular individual to be generated.

\paragraph{Revision loop.}
The model gathers photographs, records the profile, then writes a first description and up to three revisions. A call whose output fails validation is returned with the error, up to five attempts; a prompt that exhausts them ends with no plant. If the last allowed round changed the description, the model sees that version once more and keeps or reverts it, so that the result is always a plant it has seen. Before a plant is meshed, the faces it needs are estimated from its growth trace, one organ of each kind meshed and counted; a plant estimated above 25 million faces is returned with the organs that take them. The population check grows individuals 1 to 100 and fails a description when more than 5\% of them disagree with the profile in any trait; it also reports how many miss a count target. The individuals measured in \cref{sec:exp:population} are 101 to 200, which no check grows.

\paragraph{Programmatic and visual feedback.}
For the reviewed individual, trait checks read the growth trace and compare its states and ranges with the profile. A discrepancy identifies the expected state, the generated state, and the fields of the description that control it. Additional checks cover count and height targets, organ crossings, organs disconnected from the plant or entering the soil, incompatible relative organ sizes, and whether the requested support yields a standing or settled plant. Any stabilizing stiffness correction is reported to the agent.

The visual review presents the measurements and failures with whole-plant renderings and close-ups of a leaf, flower, and inflorescence beside the selected photographs. Its instructions ask the agent to review habit, stems, leaves, inflorescences, flowers, fruit, and overall size. For each discrepancy, the agent states what the organ should be and the evidence for it, what the generated plant shows, and the fields that would correct it, or the feature the vocabulary cannot express. Revisions set individual fields or replace an archetype, target, or support entry instead of rewriting the full description. The same review loop can accept a person's comment in place of the agent's visual assessment.

\paragraph{Growth.}

Each shoot is grown through node-wise apical decisions: vegetative growth adds phytomers, a reproductive apex terminates the shoot in an inflorescence, and a dormant apex stops extension. Axillary buds develop into shoots of their referenced archetypes where their node bands and sampled fates allow. Remaining reproductive buds and apices bear inflorescences, whose positions resolve to fruit, open flowers, or buds. The trace records the resulting shoots and organs, their connectivity, sampled traits, and positions.

Random choices are seeded by the individual's seed and the node's position in the plant. The same description and seed reproduce the same trace. Choices at unrelated structural positions remain unchanged when another part of the description is edited, provided those positions are preserved. Explicit organ counts are allocated to the buds and inflorescence positions the description permits; a target that cannot be placed is reported. A range of counts is fitted over the population when 90\% or more of seeds 0 to 99 would fall at one end of it: one shared factor adjusts counts across individuals so that typical individuals lie within the range while retaining their spread. A main stem that stops early leaves a plant short of its height whatever the scaling, so the main stem takes no stop draw before the least number of nodes that lets the plant reach its height. A shoot sheds the leaves of its lowest nodes only once everything in their axils has grown.

\paragraph{Engine and compiler.}
Growing an individual through the growth stage alone takes a few milliseconds. Building and checking one individual, with its geometry, physics and export, takes about a minute for most plants and needs no model call. The additional-individual times of \cref{tab:models} are medians over seeds 101 to 103 of every final description, from growth to the compiled asset, with one NVIDIA RTX 6000 Ada GPU and four CPU threads; an agent's time is the wall-clock duration of its run. Geometric queries (whether a point lies inside an organ, how far it is from one) run on a GPU. Each organ is meshed with as many faces as its size needs. The height fit scales every internode by one factor in $[0.5, 2]$. Crossings deeper than 3\% of the smaller organ are cleared where an organ can be turned or swung by up to 32 degrees at its attachment without new crossings; flowers and fruit in a crowded inflorescence that cannot clear fan out around their attachments, and a fruit that would reach into the soil swings up about its pole. Bracts that enclose an inflorescence, such as a maize ear's husks, are laid around the outline of its flowers and fruit. A stem hinge's length $l$ weights the stretch of stem on either side so that the hinge chain matches the continuous beam under a tip load. A leaf bends at its base as a beam of its petiole and then its midrib, loaded at the blade's centroid, and twists on its petiole; a fruit or inflorescence stalk bends as a cantilever, $k=3EI/L$. Fresh tissue weighs 900\,kg\,m$^{-3}$, and damping is 0.1 of critical. A fruit joint breaks at the pull-off force, and at the same force applied at the fruit's radius as a torque. The URDF that accompanies the scene holds the same bodies and joints, without the springs and break forces.

\paragraph{Joint stiffness and gravity preload.}
A stem is discretized into links at its internodes, with two bending rotations and a twisting rotation at a node. Its bending stiffness is
\begin{equation}
  k_j=\frac{E_j I_j}{l_j}, \quad I_j=\frac{\pi r_j^4}{4},
\end{equation}
where $E_j$ is the stem modulus, $r_j$ is the local radius, and $l_j$ is the effective length described above~\cite{niklas1992biomechanics}. A twisting hinge has stiffness $k_j/\tau_j$, where $\tau_j$ is the specified ratio of twisting to bending compliance~\cite{vogel1992twist}. Stem radii follow the supported leaf area; at a branching point, the cross-sections of the outgoing stems sum to that of the incoming stem~\cite{shinozaki1964pipe,eloy2011leonardo}.

The drawn pose represents a plant under its own weight. For joint angles $\boldsymbol{\theta}$ and drawn angles $\boldsymbol{\theta}_{d}$, let $U_g$ be the gravitational potential and $\mathbf K$ the diagonal spring-stiffness matrix. Spring preload balances the generalized gravitational load $\nabla U_g(\boldsymbol{\theta}_{d})$ at the drawn pose; equivalently, spring rest angles are offset to provide this balancing torque. Static balance alone does not ensure stability. With
\begin{equation}
  \mathbf H_g=\nabla^2 U_g(\boldsymbol{\theta}_{d}), \quad
  \mathbf K_{\mathrm{eff}}=\mathbf K+\mathbf H_g,
\end{equation}
the pose is locally stable when $\mathbf K_{\mathrm{eff}}$ is positive definite. Its stability margin is
\begin{equation}
  m=\frac{\lambda_{\min}(\mathbf K_{\mathrm{eff}})}
  {\mathbf v^\top \mathbf K\mathbf v},
\end{equation}
where $\mathbf v$ is a unit eigenvector for the smallest eigenvalue. The support-specific correction and settling procedures below use this criterion.

\paragraph{Physics export.}
The OpenUSD scene retains organ identities and the attachment hierarchy together with geometry, rigid-body masses and inertias, revolute joints, spring stiffnesses, damping, preloaded rest angles, joint limits, collision shapes, and a root fixed to the ground~\cite{openusd}. Fruit attachments carry their specified break-force and break-torque thresholds. Support constraints are exported with the plant; the simulator-specific collision and breakable-joint settings are described under \emph{Simulation} below.

\paragraph{Support.}
A self-supporting plant whose drawn pose is unstable has the hinges of its weakest direction stiffened together, round after round and each by at most ten times its own stiffness, until a margin of 0.1 remains; the model is told when this happens, by how much, and when even that does not hold the plant up. A staked plant's internode joints are fixed; the stake stands in the widest gap its leaves leave, with a tie every 20\,cm, and leaves and inflorescences are placed clear of it. The stake carries no load: it is welded to the stem at the lowest tie, and the ties are geometry. A twining stem is held to its pole in the same way. A sprawling plant's stems are not stiffened. The plant settles in steps from the balanced drawn pose, and from any later saddle, along a direction in which the energy curves down, until no step lowers the energy; the ground pushes back at the corners of each mesh's convex hull, as a simulator's collision would. The settled angles become the joints' rest pose, the springs keep the drawn pose's preload, and the ground carries the rest of the weight. Where the stems fold, each leaf is turned back to face as it did before settling, and the plant settles again with the stems' twists held, so that a stem lying on the ground does not roll under its leaves. A sprawling plant that holds itself up is fitted to its height like any plant; one that falls is as tall as it rests.

\paragraph{Baselines.}
Articraft runs from its released code; to reach a model served on our own GPUs, its OpenRouter provider is pointed at our server and passes the images its harness renders on to the model, as its other providers do. Codex and DeepSeek Harness run in the same container, and both search through a tool connected to the self-hosted search engine; Codex's own search tool is turned off, so that every system that searches uses the same engine. FloraForge has up to five attempts at a valid answer, as our method has. 3D-GPT, whose parsers read each function call from a single line, runs on GPT-6-astra only. The coding agents have four hours a prompt (Codex six on Qwen3.8-27B); a run stopped at its limit is scored on the asset it left in its workspace. On both self-hosted base LLMs the server holds every system's tool-call arguments to their schemas: for Qwen3.8-27B through vLLM's tool parser, and for Gemma~4~31B, whose vLLM parser lacks this constraint, through a tool-parser plugin of ours that adds it.

\paragraph{Simulation.}
MuJoCo runs through Articraft's own translation of the OpenUSD scene at its 2\,ms step, with our springs as joint stiffness and rest angles and the organs colliding with the ground but not with each other. Each hinge is given rotor inertia in proportion to its stiffness, with its damping raised to keep its damping ratio, which slows its fastest motion and changes neither a pose's balance nor the stiffness a load meets; a test that diverges is run again at 0.5\,ms, which one nudge needed. Isaac Sim~4.5 runs PhysX on the CPU with the PGS solver at 1000\,Hz; its default is TGS at 60\,Hz. PhysX fails on an articulation of more than 255 links, so the export keeps at most 255 bodies in the plant's articulation and joins the others by ordinary joints. Fruit joints that can break are also kept outside it, because PhysX ignores break forces inside an articulation. For the regrouped runs, the bodies are divided at load time into several articulations of at most 255 links, cut where the largest branches attach, and the file is not changed.

The push test's load moves the main stem's hinge chain by 1\% of the stem's length, ramped up over 2\,s and held, the movement averaged over the last 5 of 20\,s. It is applied without gravity or preload and compared with the file's hinge chain and with the stem as a continuous tapered beam, and again with both and compared with the prediction from $\mathbf K_{\mathrm{eff}}$. The nudge starts every hinge 0.05\,rad off its drawn angle (signs from a fixed seed) and runs 15\,s; a plant has returned when its largest hinge deviation stays under a tenth of its start, is returning when the deviation ends under its start, and has folded when it ends over it. In Isaac Sim, plants are held for 3\,s from the drawn pose, and five fruit joints of each fruiting plant, drawn with a fixed seed, are pulled along their stalks by a force pair ramped to three times the break force. In \cref{fig:supp:push}, the push is ten times the test's load, ramped up over 1\,s, held to 3\,s and released; the pick ramps to 1.5 times the break force over 1.5\,s, and with no pull no fruit of that plant comes off.

The wind of the supplementary video drags each rigid body through the area its surface shows to the flow: along each principal direction of that area, the force is $\tfrac{1}{2}\rho C_d A v^2$ with $\rho=1.2$\,kg\,m$^{-3}$ and $C_d=1$, applied at the area's center.

\end{document}